\documentclass{article}

\PassOptionsToPackage{numbers, sort&compress, comma, square}{natbib}
\usepackage[preprint]{neurips_2024}

\usepackage{natbib}

\usepackage[utf8]{inputenc} 
\usepackage[T1]{fontenc}    
\usepackage{hyperref}       
\usepackage{url}            
\usepackage{booktabs}       
\usepackage{amsfonts}       
\usepackage{nicefrac}       
\usepackage{microtype}      
\usepackage{xcolor}  
\usepackage{enumitem}
\usepackage{url}
\usepackage{xspace}
\usepackage{multirow}
\usepackage{amsmath}
\usepackage{subcaption}
\usepackage{graphicx}
\usepackage{colortbl}
\usepackage{caption}
\usepackage{tabularx}
\usepackage{wrapfig}

\newcommand{\shortname}{RAGFormer\xspace}

\title{RAGFormer: Learning Semantic Attributes and Topological Structure for Fraud Detection}

\newcommand{\equalcontrib}{\footnotemark[1]}
\newcommand{\corresponding}{\footnotemark[2]}

\author{%
    \textbf{Haolin Li}$^{1}\equalcontrib$ \quad
	\textbf{Shuyang jiang}$^{1}\equalcontrib$ \quad
	\textbf{Lifeng Zhang}$^{1,2,3}\equalcontrib$ \quad \\
	\textbf{Siyuan Du}$^{1}$ \quad
	\textbf{Guangnan Ye}$^{1,2}\corresponding$ \quad
    \textbf{Hongfeng Chai}$^{1,2}$
	\vspace{2mm} \\
	$^{1}$School of Computer Science, Fudan University \quad
	$^{2}$Institute of FinTech, Fudan University \\
	$^{3}$Shanghai ChinaPay e-payment Service Co.,Ltd \quad
	\vspace{2mm} \\
	\textnormal{\{23110240025, 23110240021\}@m.fudan.edu.cn} \quad
	\textnormal{zhanglifeng@chinapay.com} \quad \\
	\textnormal{23110240011@m.fudan.edu.cn} \quad
	\textnormal{\{yegn, hfchai\}@fudan.edu.cn} 
}

\begin{document}

\maketitle

\renewcommand{\thefootnote}{\fnsymbol{footnote}}
\footnotetext[1]{Equal contribution.}
\footnotetext[2]{Corresponding author.}

\begin{abstract}
 Fraud detection remains a challenging task due to the complex and deceptive nature of fraudulent activities.
 Current approaches primarily concentrate on learning only one perspective of the graph: either the topological structure of the graph or the attributes of individual nodes. 
 However, we conduct empirical studies to reveal that these two types of features, while nearly orthogonal, are each independently effective.
 As a result, previous methods can not fully capture the comprehensive characteristics of the fraud graph.
  To address this dilemma, we present a novel framework called \underline{R}elation-\underline{A}ware \underline{G}NN with trans\underline{Former}~(\shortname) which simultaneously embeds both semantic and topological features into a target node.
  The simple yet effective network consists of a semantic encoder, a topology encoder, and an attention fusion module. 
  The semantic encoder utilizes Transformer to learn semantic features and node interactions across different relations. 
  We introduce Relation-Aware GNN as the topology encoder to learn topological features and node interactions within each relation. 
  These two complementary features are interleaved through an attention fusion module to support prediction by both orthogonal features.
  Extensive experiments on two popular public datasets demonstrate that \shortname achieves state-of-the-art performance.
  The significant improvement of \shortname in an industrial credit card fraud detection dataset further validates the applicability of our method in real-world business scenarios.
\end{abstract}

\section{Introduction}


As the Internet continues to thrive, a plethora of fraudulent activities has increasingly emerged. Consequently, fraud detection has garnered more attention in various fields such as e-commerce~\citep{liu2020fraud,hu2020loan,zhang2022efraudcom}, review management~\citep{rayana2015collective,dou2020enhancing,nilizadeh2019think}, and social networking~\citep{yuan2019detecting,chen2020phishing}. 
Leveraging the capability of graph structures in representing the topological structure of data, graph-based neural networks have emerged as the preeminent methodology in tackling fraud detection challenges.
Typically, fraud detection datasets are organized into multi-relation graphs, in which the various relations between entities are represented as distinct types of edges. 
For instance, in credit card fraud detection, users are represented as nodes and two primary types of relations can be delineated: (1) users engaging in transactions with the same merchant, and (2) users conducting transactions in the same manner. 


To model the relations between nodes, many GNN-based fraud detectors have been proposed~\citep{dou2020enhancing,zang2023don,wang2023removing,yu2023group}. 
GNN employs message passing to iteratively aggregate information from neighborhoods, thereby effectively capturing the relations between nodes.
Consequently, GNN is adept at learning the topological structure of the multi-relation graph. 
Relying on the powerful sequence modeling capabilities of Transformer~\citep{vaswani2017attention}, methods based on Transformer~\citep{liu2022user,wang2023label} have recently surpassed GNN-based methods.
With the multi-head self-attention mechanism, the Transformer encoder is capable of learning high-quality node embeddings. 
Transformer-based fraud detectors usually select the n-hop subgraph of the target node and transform it into sequence data. 
However, we observed that the performance of the Transformer-based method remains unaffected by the number of hops in the subgraph. This indicates that Transformers primarily focus on learning the intrinsic attributes of individual nodes rather than the relations between nodes. 
As illustrated in Figure~\ref{fig:sim}, our preliminary experiments further demonstrate that the features learned by GNN and Transformer are nearly orthogonal. Specifically, GNN is more focused on learning the topological structure of the multi-relation graph, whereas Transformer concentrates on learning the semantic attributes of the target nodes.
Therefore, both types of methods—whether based on GNN or Transformer—are inherently limited as they each excel in learning only one aspect of the graph.
Inspired by the above analysis, we propose a novel framework to capture both semantic and topological node features in a unified pipeline.
Our network consists of three modules: the semantic encoder, the topology encoder, and the attention fusion module.
The semantic encoder utilizes Transformer to learn semantic features and 
integrates an additional cross-relation aggregation layer between any two successive Transformer layers to enhance relational interactions.
The topology encoder (Relation-Aware GNN) employs multiple relation-specific GNNs (typically GCN~\citep{velivckovic2017graph}) to learn topological features for the target node. 
It directly models each relation with a corresponding GNN and learns multi-relation embeddings in an end-to-end manner.
Finally, the semantic and topological features are merged by the feature fusion module with the self-attention mechanism. 
We follow the design of Transformer~\citep{vaswani2017attention} to add a residual connection after the self-attention to prevent the model from degrading to a vanilla GNN.
In a nutshell, we conclude our contributions as follows:
\begin{itemize}[leftmargin=*]
    \item Our research reveals that both Transformer-based and GNN-based methods can learn only one aspect of the graph. Transformers tend to overlook topological information, whereas GNNs struggle to learn high-quality semantic information. 
    The embeddings they learn are almost orthogonal with low correlation, thus complementing each other effectively without introducing any redundancy.
    \item We propose \shortname, a unified framework that integrates three modules to learn both semantic and topological embeddings among a heterogeneous graph. The semantic encoder captures discriminative semantic features, while the topology encoder learns topological features in an end-to-end manner. The attention fusion module then efficiently merges the semantic and topological features together, leading to a comprehensive representation of the multi-relation graph.
    \item Extensive experiments on 
    two public opinion fraud datasets and an industrial financial fraud dataset demonstrate the superiority of our method.
    \shortname shows substantial improvement on large-scale datasets especially, with up to 12.08\% improvement over state-of-the-art methods.
\end{itemize}



\section{Related Work}

\textbf{GNN-based fraud detectors.} 
GNN-based methods have demonstrated outstanding performance in fraud detection. Various approaches have been proposed to address specific challenges in this domain.
CARE-GNN~\citep{dou2020enhancing} employed novel neighbor filtering techniques to overcome the camouflage problem. 
$H^{2}$-FDetector~\citep{shi2022h2} designed distinct aggregation strategies for homophilic and heterophilic neighbors.
GTAN~\citep{xiang2023semi} exploited an attribute-driven gated temporal attention network for credit card fraud detection.
Previous GNN-based methods typically transformed multi-relation heterogeneous graphs into homogeneous graphs~\citep{wang2019semi,wang2019fdgars,liu2020alleviating}, or developed various strategies for learning directly from heterogeneous graphs~\citep{liu2018heterogeneous,zhong2020financial,li2021live,liu2021intention,zang2023don}. However, most of these methods used a single GNN to model multi-relation graphs. In contrast, we design our topology encoder with a Relation-Aware GNN architecture that applies a specific GNN for each type of relation, enabling end-to-end modeling of multi-relation graphs. 
In addition to the aforementioned approaches that overlook relations, there are other methods, such as R-GCN~\citep{schlichtkrull2018modeling}, that take into account multiple types of relations. R-GCN utilized a single model with multiple weight matrices, where each matrix corresponded to a type of relation, and these matrices were jointly aggregated. In contrast, our Relation-Aware GNN employs multiple distinct GNNs, with each model aggregating messages separately.

\textbf{Transformer-based fraud detectors.} 
Beyond the aforementioned GNN-based detectors, recent works~\citep{liu2021anomaly,liu2022user,wang2023label} have made early attempts at using Transformer for fraud detection. These approaches converted heterogeneous graphs into sequential data and introduced additional encodings to incorporate temporal or relational information into the Transformer model. UB-PTM~\citep{liu2022user} learned the characteristics of online fraudulent activities from the three agent tasks: action, intention, and sequence. GAGA~\citep{wang2023label} introduced additional hop encoding, label encoding, and relation encoding to incorporate structural and labeling information into the Transformer model. Compared with these Transformer-based approaches, our semantic encoder incorporates cross-relation aggregation between adjacent Transformer layers to seamlessly integrate information across different relations.

\begin{figure}[tbp]
    \centering  
    \begin{subfigure}[c]{0.45\textwidth}
         \centering
        \includegraphics[width=\textwidth]{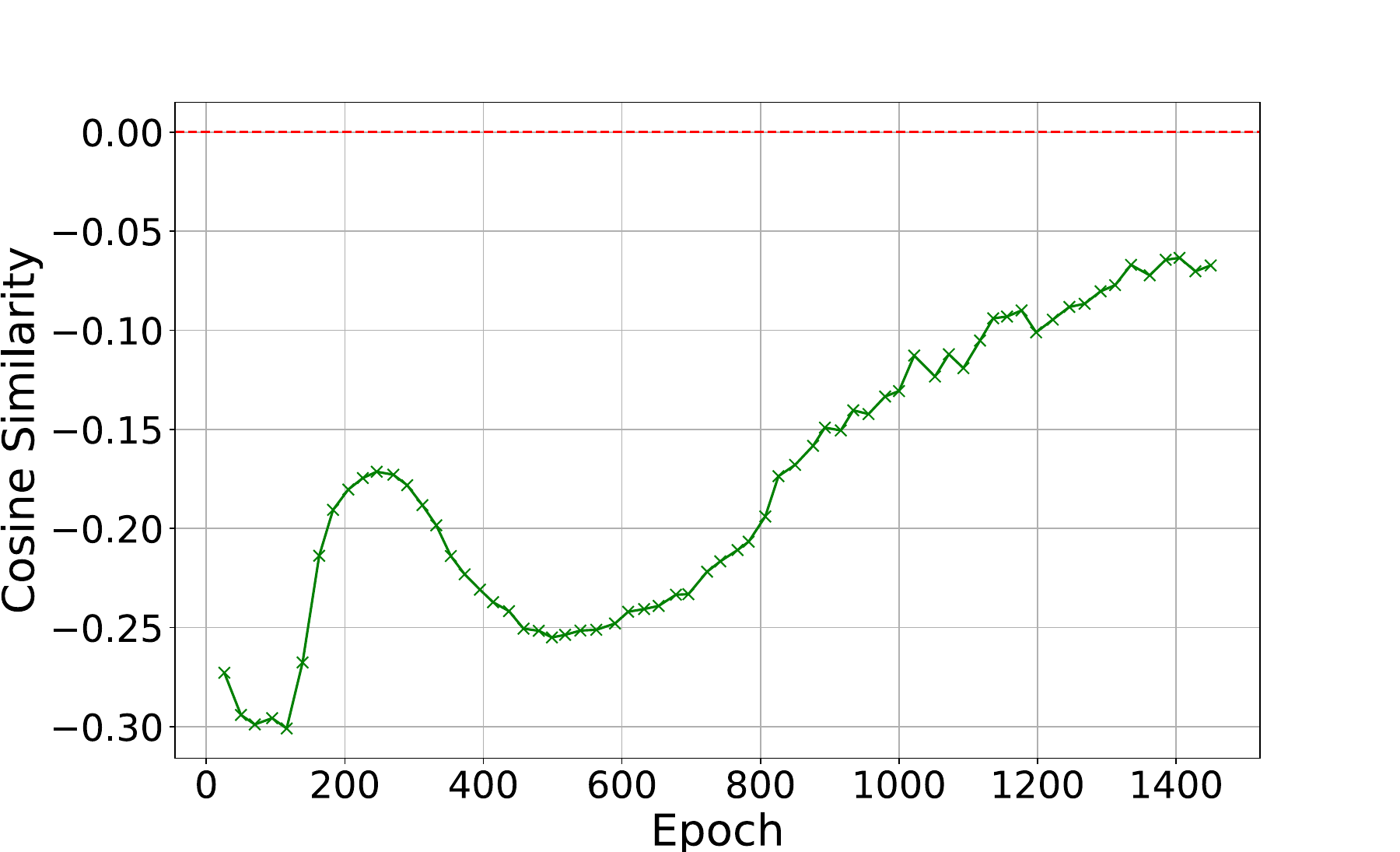}
        \caption{Cosine Similarity on YelpChi}
         \label{fig:sim-y-cos}
     \end{subfigure}
     \begin{subfigure}[c]{0.45\textwidth}
         \centering
        \includegraphics[width=\textwidth]{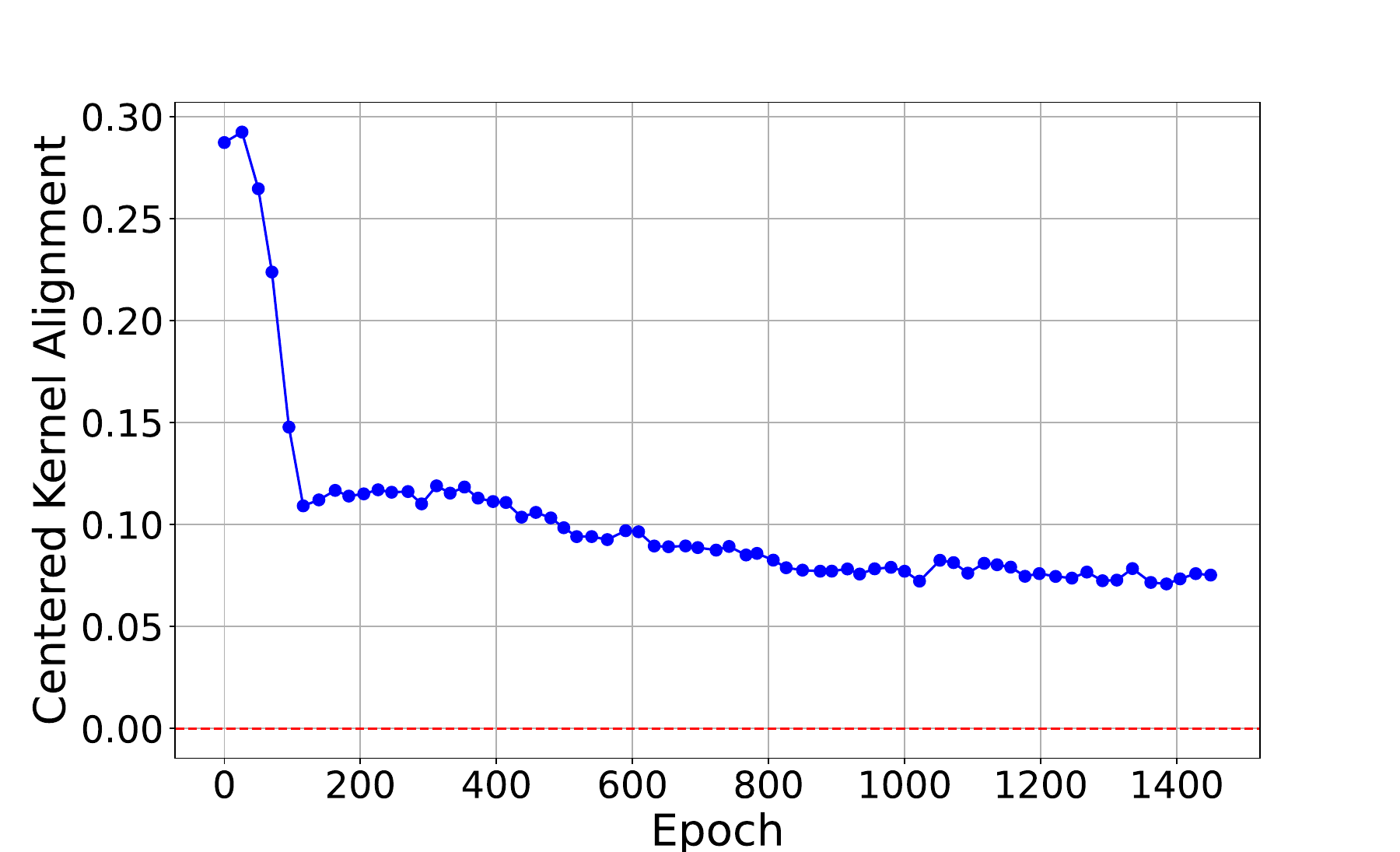}
        \caption{CKA on YelpChi}
         \label{fig:sim-y-cka}
     \end{subfigure}
\caption{Similarity between features learned by our semantic encoder(Transformer) and topology encoder(GNN). 
A value of 0, indicated by the red line, signifies that the features are orthogonal.}
\label{fig:sim}
\end{figure}

\section{Preliminaries}
In section 3.1, we define the multi-relation graph. In section 3.2, we briefly discuss the limitations of the Transformer-based method and analyze the similarity between embeddings learned by Transformer and GNN through preliminary experiments.

\subsection{Notations}
\textbf{Multi-relation Graph.} We define a multi-relation graph as $\mathcal{G}=\{\mathcal{V,H,A,Y}\}$. $\mathcal{V}=\{v_{1},...,v_{n}\}$ is the set of nodes and $n$ is the number of nodes. $\mathcal{H}=\{H_{1},...,H_{n}\}$ is the set of node features. For each node $v_{i}$, $h_{i} \in \mathbb{R}^{k}$ is the corresponding k-dimension feature vector. $\mathcal{A}=\{A^{1},...,A^{R}\}$ is the set of adjacent matrix and $R$ is the number of relations. $a_{i,j}^{r} \in A^{r}$ represents the edge between node $v_{i}$ and $v_{j}$ under relation $r$. $\mathcal{Y}=\{y_{1},...,y_{n}\}$ is the set of labels, where $y_{i} \in \{0, 1\}$. $y_{i}=0$ denotes that node $i$ is a benign node and $y_{i}=1$ denotes that node $i$ is a fraud node.
$\mathcal{G}$ can be divided into relation-specific subgraphs 
$\mathcal{G}_{1},...,\mathcal{G}_{R}$, where $\mathcal{G}_{i}=\{\mathcal{V},\mathcal{H},A^{i},\mathcal{Y}\}$.


\begin{wraptable}{r}{0.35\textwidth}
\vspace{-1em}
    \centering
    \caption{Preliminary experiment on YelpChi. 
    }
    \label{tab:preliminary}
    \resizebox{0.35\textwidth}{!}{
    \begin{tabular}{c|c|c|c}
        \toprule
        \multicolumn{1}{c|}{\multirow{2}{*}{\textbf{Subgraph}}} &\multicolumn{3}{c}{\textbf{YelpChi}} \\
        \cmidrule{2-4} \multicolumn{1}{c|}{}& \textbf{AUC} & \textbf{AP} & \textbf{F1} \\
        \midrule
        1-hop & 0.9432 & 0.7954 & 0.8277 \\
        2-hop & 0.9435 & 0.7968 & 0.8299 \\
        3-hop & 0.9437 & 0.7937 & 0.8303 \\
        4-hop & 0.9437 & 0.7972 & 0.8347\\
        5-hop & 0.9417 & 0.7951 & 0.8369\\
        \bottomrule
    \end{tabular}}
    \vspace{-1em}
\end{wraptable}

\subsection{Motivation}
\label{motivation}
The most advanced Transformer-based fraud detector,  GAGA~\citep{wang2023label} was proposed to model graphs under low-homophily settings.
GAGA extracts a 2-hop subgraph $\mathcal{G}_{sub}$, from the original graph $\mathcal{G}$ for each node under different relations.
Each subgraph is flattened into a sequence and input to the Transformer.

While GAGA~\citep{wang2023label} has achieved state-of-the-art performance on fraud detection task, it still exhibits certain limitations. Specifically, GAGA can not learn the topological
structure around the target node, leaving this method unable to mine relational information between neighbors. To prove this, we conducted a preliminary experiment on the YelpChi~\citep{rayana2015collective} dataset to explore how varying the number of hops within the subgraph affects the performance. Assuming that GAGA can capture the topology of the data, then increasing the number of hops should expand the model's receptive field, thereby improving the performance.

As illustrated in Table~\ref{tab:preliminary}, expanding the number of hops from one to five fails to enhance GAGA's performance. In fact, the number of hops in the subgraph does not influence the learning process of the Transformer. 
This indicates that GAGA is unable to capture the topological structure of the data and is limited to learning node attributes. 
Its effectiveness primarily stems from the powerful sequence encoding ability of Transformer. 
Graph Neural Networks have demonstrated their capability in learning discriminative node representations on large-scale multi-hop graphs, with their ability to learn topological structures fully verified. Our intuition is that embeddings learned by GNNs and those learned by Transformers focus on different aspects of the data, constituting a certain complementarity. Based on this insight, we design our \shortname by combining GNN and Transformer together.

To validate our intuition, we evaluate the similarity between features learned by our semantic encoder and topology encoder on two datasets, YelpChi~\citep{rayana2015collective} and Amazon~\citep{mcauley2013amateurs}. We adopt cosine similarity and Centered Kernel Alignment (CKA)~\citep{kornblith2019similarity} as metrics and illustrate their changes over training epochs in Fig.~\ref{fig:sim}. 
CKA measures the similarity between two sets of features, with values close to 1 indicating strong similarity and values near 0 indicating little or no similarity. 
On YelpChi, as the number of training epochs increases, both the cosine similarity and CKA score between the semantic features and topology features gradually approach zero, indicating that the features are nearly orthogonal with low correlation. 
The representations learned by the semantic encoder and the topology encoder are nearly orthogonal, yet both are effective for fraud detection. 
This confirms our intuition that Transformer and GNN indeed focus on different aspects of the graph, thereby suggesting that combining the two could further boost performance.
More results on the Amazon dataset are provided in Appendix~\ref{pre-app}.

\begin{figure}[tbp]
    \centering
    \includegraphics[width=0.98\textwidth]{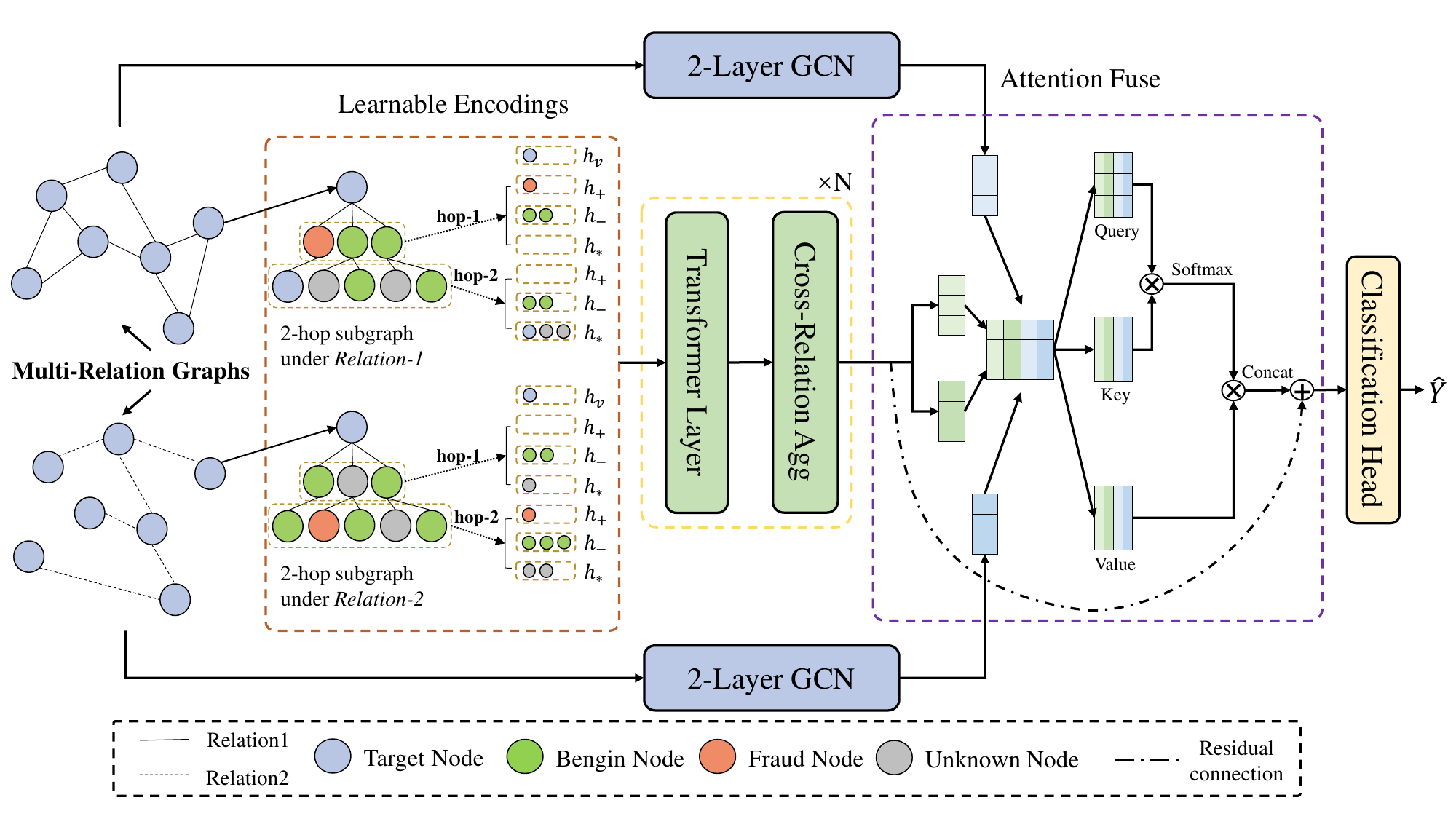}
    \caption{The pipeline of
    our \shortname method that learns both semantic features and topological features. 
    The complete multi-relation graph $\mathcal{G}$ is input into the semantic encoder. The topology encoder contains $R$ independent GNNs. We extract the relation-specific subgraphs $\mathcal{G}_{1},...,\mathcal{G}_{R}$ as the input of the corresponding GNN. The semantic features $X_{sem}$ and the topological features $X_{gcn}$  are then input into the feature fusion module to get the final embeddings $X_{fused}$. A classification head, with the Multi-Layer Perceptron (MLP) architecture, takes $X_{fused}$ as the input and outputs the classification probabilities $\hat{\mathcal{Y}}$. We set the relation number $R$ to 2 within the graph for simplicity.} 
    \label{fig:method}
\end{figure}

\section{Method}
In this section, we present the proposed method \shortname. Fig.~\ref{fig:method}  illustrates the overall framework.
We introduce detailed information of \shortname in three parts: (a) the semantic encoder that learns semantic features through the Transformer with cross-relation aggregation layers; (b) the topology encoder that embeds topological features through Relation-Aware GNN; and (c) the feature fusion module that combines semantic and topological features in an attentive way.


\subsection{Semantic Encoder}
We follow the preprocessing procedure in \S\ref{motivation} to extract the two-hop subgraph and convert it into a sequence.
Each node in the extracted subgraph has three types of information: hop distance from the target node, label information from training sets, and relation types.
Therefore, to distinguish the identities of different nodes in the Transformer input, we follow GAGA~\citep{wang2023label} to use hop-embedding $\mathbf{E}_h$, group-embedding $\mathbf{E}_g$ and relation-embedding $\mathbf{E}_r$ to encode the above-mentioned features into the target node, respectively.
Specifically, the hop-embedding, group-embedding, and relation-embedding are added together to the original token embedding: $\mathbf{X} = \mathbf{X}_n+\mathbf{E}_g+\mathbf{E}_h+\mathbf{E}_r$,
where $\mathbf{X}$ is the input sequence of the Transformer encoder and $\mathbf{X}_n$ is the original subgraph node embedding.

After embedding the target node, we use a Transformer~\citep{vaswani2017attention} encoder to learn the semantic features.
To fuse different relation features fully, we add a cross-relation aggregation layer after each Transformer layer.
Formally, for each output of $l$-th layer of Transformer encoder $\mathbf{X}_{o;l}$, we introduce an aggregation layer to fuse different relation features into the target node $\boldsymbol{X}_{t;l}$:
\begin{align}
    \boldsymbol{X}_{t;l} &= [\boldsymbol{X}_{ot;l}^{1},\boldsymbol{X}_{ot;l}^{2}\cdots,\boldsymbol{X}_{ot;l}^{R}] \\
    \boldsymbol{X}_{t;l+1} &= W\boldsymbol{X}_{t;l},
\end{align}
where $[\cdot]$ is the concatenation operator, superscripts $1,2,\cdots,R$ refer to relation indexes and $W$ is a learnable parameter for dimension alignment.
The cross-relation aggregation layers only introduce a few parameters, but bring further performance gain.

\subsection{Topology Encoder}
The semantic encoder is powerful at learning the intrinsic features of the target node.
However, it lacks the ability to extract information from the neighbors of the target node, thus failing to learn the relation between adjacent nodes.
To address this issue, we add the topology encoder aside from the semantic encoder. We propose Relation-Aware GNN as the architecture of the topology encoder. The Relation-Aware GNN contains multiple 2-layer GCNs, each takes a relation-specific subgraph $\mathcal{G}_{r}$ as the input.
For each relation, the complete graph of the current batch is input into the corresponding GCN to obtain the global representation for each node. Concretely, for relation $r$, the output of the $l$-th layer of GCN is: 
\begin{align}
\boldsymbol{X}_{gcn}^{l,r}=\sigma(\Tilde{D}^{-\frac{1}{2}}\Tilde{A}^r\Tilde{D}^{-\frac{1}{2}}H^{l,r}W^{l,r}),
\end{align}
where $\Tilde{A}^r\in\mathbb{R}^{N\times N}$ is the adjacent matrix under relation $r$, $H^{l}$ is the graph feature and $\Tilde{D}$ is the degree matrix of $\Tilde{A}^r$. The output of the topology encoder is $\boldsymbol{X}_{gcn}=\{\boldsymbol{X}_{gcn}^{1},...,\boldsymbol{X}_{gcn}^{R}\}$, where $\boldsymbol{X}_{gcn}^{r}$ is the last layer output of the GNN corresponding to relation $r$.

\subsection{Feature Fusion Attentive Module}
After obtaining the topological features learned by the GNN encoder $X_{gcn}$ and the semantic features $X_{sem}$ learned by the Transformer encoder, 
we fuse the semantic feature and the topological feature using the self-attention~\citep{vaswani2017attention}.
Specifically, the semantic node feature $\boldsymbol{X}_{sem}=\{\boldsymbol{X}_{sem}^{1},...,\boldsymbol{X}_{sem}^{R}\}$, where $\boldsymbol{X}_{sem}^{r}\in\mathbb{R}^{n\times d}$ is the embeddings of relation $r$, $n$ is the batch size and $d$ is the embedding dimension. 
We first stack all node embeddings in $X_{gcn}$ and $X_{sem}$ into a sequence $X_{seq}$ of length $2R$.
And then employ a self-attention layer followed by a residual connection to obtain the fused feature $X_{fused}$:
\begin{align}
    \boldsymbol{\hat{X}}_{seq} = \mathrm{SelfAttention}(\boldsymbol{X}_{seq}) \\
    \boldsymbol{X}_{fused} = \boldsymbol{X}_{sem}+\mathrm{Linear}(\boldsymbol{\hat{X}}_{seq}),
\end{align}
where $\mathrm{Linear}()$ is a linear layer to align the embedding dimension and $\boldsymbol{X}_{fused}$ is the final fused representation of the target nodes. 
We use residual connection~\citep{he2016identity} to guarantee that the semantic features can mix with topological features without disturbing its learning process.

$\boldsymbol{X}_{fused}$ is then concatenated and input to the classification head to get the final prediction.
The attention fusion module can efficiently integrate topological features into semantic features, and accelerate the convergence of the model.

\subsection{Training}
The final learning procedure is to perform cross-entropy loss against ground truth labels:
\begin{align}
    \mathcal{L} = \sum_{i\in\mathcal{V}}[y_i\log p_i+(1-y_i)\log p_i] \quad
    p_i &= \mathrm{sigmoid}(\mathrm{MLP}(\boldsymbol{X}_{fused})),
\end{align}
where MLP is the classification head, $p_i$ is the predicted probability and $y_i$ is the ground truth label for node $i$.

\section{Experiments}
To evaluate \shortname's effectiveness, we conducted experiments on two real-world opinion fraud datasets: YelpChi view~\citep{rayana2015collective} and Amazon review~\citep{mcauley2013amateurs} and an industrial financial fraud dataset Pay. Our code is available at \url{https://github.com/tdlhl/RAGFormer}.


\subsection{Experimental Setup}
\textbf{Datasets.}
We utilize the YelpChi~\citep{rayana2015collective} and Amazon~\citep{mcauley2013amateurs} datasets to evaluate the performance of \shortname on opinion fraud detection. 
To verify the effectiveness of \shortname in detecting financial fraud, we curated a real-world industrial dataset named Pay. Pay is designed for credit card fraud detection and comprises the transaction flow of users who used their credit cards for transactions over a six-month period in 2023, along with detailed attributes of the corresponding users.
The detailed statistics of the three multi-relation datasets can be found in Appendix~\ref{dataset-app}.

\textbf{Baselines.}
We compare our \shortname with three categories of methods as follows: (1).Classical GNN models, including GCN~\citep{kipf2016semi}, GAT~\citep{velivckovic2017graph}, GraphSAGE~\citep{hamilton2017inductive}, GraphSAINT~\citep{zeng2019graphsaint} and heterogeneous model HAN~\citep{han2019}. (2).GNN-based fraud detectors, including GraphConsis~\citep{liu2020alleviating}, CARE-GNN~\citep{dou2020enhancing}, PC-GNN~\citep{liu2021pick}, RioGNN~\citep{peng2021reinforced}, FRAUDRE~\citep{zhang2021fraudre}, H$^2$-FDetector~\citep{shi2022h2} and COFRAUD~\citep{zang2023don}. (3).Transformer-based fraud detector GAGA~\citep{wang2023label}, which is the state-of-the-art method.

\textbf{Implementation Details.}
Following previous works~\citep{dou2020enhancing,liu2021pick,wang2023label}, we set the training percentage to be 40\% on all datasets. Note that classical GNN models can not handle multi-relation graphs directly, thus we converted the multi-relation graph into a homogeneous graph for these models.
We adopted three widely-used evaluation metrics~\citep{ling2003auc,yang1999evaluation} to evaluate the performance of all methods, including AUC, AP, and F1-macro. 
We used Adam~\citep{kingma2014adam} with a learning rate of 0.001 and weight decay of 0.0001 as the optimizer for all methods. 
Our code was implemented by Pytorch~\citep{paszke2019pytorch} and DGL~\citep{wang2019deep}.

\begin{table*}[tbp]
    \caption{Opinion fraud detection performance on YelpChi and Amazon.
    The \textbf{boldface} means the best results.
    Our method is highlighted with shading.
    Some of the scores are missing due to the corresponding papers not reporting or providing open-source code.
    }
    \label{tab:performance}
    \centering
    \fontsize{8}{8}\selectfont
	\setlength\tabcolsep{3pt}
    \begin{tabular}{c|c|c|c|c|c|c}
        \toprule
        \multicolumn{1}{c|}{\multirow{2}{*}{\textbf{Methods}}} &\multicolumn{3}{c|}{\textbf{YelpChi}} &\multicolumn{3}{c}{\textbf{Amazon}} \\
        \cmidrule{2-7}\multicolumn{1}{c|}{}& \textbf{AUC} & \textbf{AP} & \textbf{F1-macro} & \textbf{AUC} & \textbf{AP} & \textbf{F1-macro} \\
        \midrule
        GCN~\citep{kipf2016semi} & 0.5924\fontsize{7}{7}\selectfont\(\pm\)0.0030 & 0.2176\fontsize{7}{7}\selectfont\(\pm\)0.0119 & 0.5072\fontsize{7}{7}\selectfont\(\pm\)0.0271& 0.8405\fontsize{7}{7}\selectfont\(\pm\)0.0075 & 0.4660\fontsize{7}{7}\selectfont\(\pm\)0.0131&  0.6985\fontsize{7}{7}\selectfont\(\pm\)0.0046 \\
        GAT~\citep{velivckovic2017graph} & 0.6796\fontsize{7}{7}\selectfont\(\pm\)0.0070 & 0.2807\fontsize{7}{7}\selectfont\(\pm\)0.0048 & 0.5773\fontsize{7}{7}\selectfont\(\pm\)0.0080 & 0.8096\fontsize{7}{7}\selectfont\(\pm\)0.0113 & 0.3082\fontsize{7}{7}\selectfont\(\pm\)0.0067 & 0.6681\fontsize{7}{7}\selectfont\(\pm\)0.0076 \\
        GraphSAGE~\citep{hamilton2017inductive} & 0.7409\fontsize{7}{7}\selectfont\(\pm\)0.0000 & 0.3258\fontsize{7}{7}\selectfont\(\pm\)0.0000 & 0.6001\fontsize{7}{7}\selectfont\(\pm\)0.0002 & 0.9172\fontsize{7}{7}\selectfont\(\pm\)0.0001 & 0.8268\fontsize{7}{7}\selectfont\(\pm\)0.0002 & 0.9029\fontsize{7}{7}\selectfont\(\pm\)0.0004 \\
        GraphSAINT~\citep{zeng2019graphsaint} & 0.7412\fontsize{7}{7}\selectfont\(\pm\)0.0143 & 0.3641\fontsize{7}{7}\selectfont\(\pm\)0.0304 & 0.5974\fontsize{7}{7}\selectfont\(\pm\)0.0728 & 0.8946\fontsize{7}{7}\selectfont\(\pm\)0.0176 & 0.7956\fontsize{7}{7}\selectfont\(\pm\)0.0091 & 0.8888\fontsize{7}{7}\selectfont\(\pm\)0.0244 \\
        HAN~\citep{han2019} & 0.8275\fontsize{7}{7}\selectfont\(\pm\)0.0000 & 0.4706\fontsize{7}{7}\selectfont\(\pm\)0.0004 & 0.6671\fontsize{7}{7}\selectfont\(\pm\)0.0012 & 0.8283\fontsize{7}{7}\selectfont\(\pm\)0.0000 & 0.4857\fontsize{7}{7}\selectfont\(\pm\)0.0015 & 0.6524\fontsize{7}{7}\selectfont\(\pm\)0.0034 \\
        \midrule
        GraphConsis~\citep{liu2020alleviating} & 0.6983\fontsize{7}{7}\selectfont\(\pm\)0.0302 & / & 0.5870\fontsize{7}{7}\selectfont\(\pm\)0.0200 & 0.8741\fontsize{7}{7}\selectfont\(\pm\)0.0334 & / & 0.7512\fontsize{7}{7}\selectfont\(\pm\)0.0325 \\ 
        CARE-GNN~\citep{dou2020enhancing} & 0.7854\fontsize{7}{7}\selectfont\(\pm\)0.0111 & 0.3972\fontsize{7}{7}\selectfont\(\pm\)0.0208 & 0.6064\fontsize{7}{7}\selectfont\(\pm\)0.0186 & 0.8823\fontsize{7}{7}\selectfont\(\pm\)0.0305 & 0.7609\fontsize{7}{7}\selectfont\(\pm\)0.0904 & 0.8592\fontsize{7}{7}\selectfont\(\pm\)0.0547 \\
        FRAUDRE~\citep{zhang2021fraudre} & 0.7588\fontsize{7}{7}\selectfont\(\pm\)0.0078 & 0.3870\fontsize{7}{7}\selectfont\(\pm\)0.0186 & 0.6421\fontsize{7}{7}\selectfont\(\pm\)0.0135 & 0.9308\fontsize{7}{7}\selectfont\(\pm\)0.0180 & 0.8433\fontsize{7}{7}\selectfont\(\pm\)0.0089 & 0.9037\fontsize{7}{7}\selectfont\(\pm\)0.0031 \\
        PC-GNN~\citep{liu2021pick} & 0.8154\fontsize{7}{7}\selectfont\(\pm\)0.0031 & 0.4797\fontsize{7}{7}\selectfont\(\pm\)0.0064 & 0.6523\fontsize{7}{7}\selectfont\(\pm\)0.0197 & 0.9489\fontsize{7}{7}\selectfont\(\pm\)0.0067 & 0.8435\fontsize{7}{7}\selectfont\(\pm\)0.0166 & 0.8897\fontsize{7}{7}\selectfont\(\pm\)0.0144 \\
        RioGNN~\citep{peng2021reinforced} & 0.8144\fontsize{7}{7}\selectfont\(\pm\)0.0050 & 0.4722\fontsize{7}{7}\selectfont\(\pm\)0.0079 & 0.6422\fontsize{7}{7}\selectfont\(\pm\)0.0233 & 0.9558\fontsize{7}{7}\selectfont\(\pm\)0.0019 & 0.8700\fontsize{7}{7}\selectfont\(\pm\)0.0044 & 0.8848\fontsize{7}{7}\selectfont\(\pm\)0.0125 \\
        H\(^2\)-FDetector~\citep{shi2022h2} &0.8892\fontsize{7}{7}\selectfont\(\pm\)0.0020 & 0.5543\fontsize{7}{7}\selectfont\(\pm\)0.0135 & 0.7345\fontsize{7}{7}\selectfont\(\pm\)0.0086 & 0.9605\fontsize{7}{7}\selectfont\(\pm\)0.0008 & 0.8494\fontsize{7}{7}\selectfont\(\pm\)0.0023 & 0.8010\fontsize{7}{7}\selectfont\(\pm\)0.0058 \\
        COFRAUD~\citep{zang2023don} & 0.9152 & / & 0.7971 & \textbf{0.9721} & / & 0.9153 \\
        \midrule
        GAGA~\citep{wang2023label} & 0.9439\fontsize{7}{7}\selectfont\(\pm\)0.0016 & 0.8014\fontsize{7}{7}\selectfont\(\pm\)0.0063 & 0.8323\fontsize{7}{7}\selectfont\(\pm\)0.0041 & 0.9629\fontsize{7}{7}\selectfont\(\pm\)0.0052 & 0.8815\fontsize{7}{7}\selectfont\(\pm\)0.0095 & 0.9133\fontsize{7}{7}\selectfont\(\pm\)0.0040\\
        \midrule
        \rowcolor{gray!10}
        \textbf{\shortname~(Ours)}  & \textbf{0.9781\fontsize{7}{7}\selectfont\(\pm\)0.0012} & \textbf{0.9222\fontsize{7}{7}\selectfont\(\pm\)0.0059} & \textbf{0.9098\fontsize{7}{7}\selectfont\(\pm\)0.0045} & 
        0.9712\fontsize{7}{7}\selectfont\(\pm\)0.0014 & \textbf{0.8936\fontsize{7}{7}\selectfont\(\pm\)0.0027} & \textbf{0.9180\fontsize{7}{7}\selectfont\(\pm\)0.0051} \\
        \bottomrule
    \end{tabular}
\end{table*}

\subsection{Quantitative Results}
Table~\ref{tab:performance} and Table~\ref{tab:Pay} present the overall performance of the proposed method \shortname and other baselines on opinion and financial fraud detection respectively. 
Table~\ref{tab:performance} shows that \shortname achieves the state-of-the-art~(SOTA) performance compared to previous methods on the YelpChi dataset, which better reflects the performance gaps between algorithms.
Specifically, \shortname surpasses the previous best method GAGA, with 12.08\% AP improvement, which demonstrates the effectiveness of merging semantic and topological features from a holistic view.
Moreover, on the relatively smaller dataset, Amazon, \shortname still maintains the lead on all metrics compared to most methods and performs comparably with COFRAUD~\citep{zang2023don}, a strong baseline that focuses on encoding multi-relation interactions.
On the financial fraud detection dataset Pay, \shortname again shows its superiority on all metrics on Table~\ref{tab:Pay}, where \shortname outperforms GAGA with $\approx$ 3\%-5\% across all metrics.
\begin{wraptable}{r}{0.5\textwidth}
    \centering
    \caption{Financial fraud detection performance on the Pay dataset. 
    Here we compare \shortname with open-sourced models, including four traditional GNN-based methods and two fraud detection oriented methods.
    }
    \label{tab:Pay}
    \resizebox{0.5\textwidth}{!}{
    \begin{tabular}{c|c|c|c}
        \toprule
        \multicolumn{1}{c|}{\multirow{2}{*}{\textbf{Methods}}} &\multicolumn{3}{c}{\textbf{Pay}} \\
        \cmidrule{2-4}\multicolumn{1}{c|}{}& \textbf{AUC} & \textbf{AP} & \textbf{F1-macro} \\
        \midrule
        GCN (ICLR'17) & 0.5170\fontsize{7}{7}\selectfont\(\pm\)0.0000 & 0.5001\fontsize{7}{7}\selectfont\(\pm\)0.0001 & 0.3708\fontsize{7}{7}\selectfont\(\pm\)0.0000 \\
        GAT (ICLR'17) & 0.5001\fontsize{7}{7}\selectfont\(\pm\)0.0070 & 0.3082\fontsize{7}{7}\selectfont\(\pm\)0.0054 & 0.3545\fontsize{7}{7}\selectfont\(\pm\)0.0000 \\
        GraphSAGE (NeurIPS'17) & 0.8588\fontsize{7}{7}\selectfont\(\pm\)0.0000 & 0.7788\fontsize{7}{7}\selectfont\(\pm\)0.0001 & 0.7739\fontsize{7}{7}\selectfont\(\pm\)0.0001 \\
        HAN (WWW'19) & 0.8553\fontsize{7}{7}\selectfont\(\pm\)0.0000 & 0.7886\fontsize{7}{7}\selectfont\(\pm\)0.0006 & 0.7442\fontsize{7}{7}\selectfont\(\pm\)0.0026 \\
        \midrule
        CARE-GNN (CIKM'20) & 0.9014\fontsize{7}{7}\selectfont\(\pm\)0.0007 & 0.8502\fontsize{7}{7}\selectfont\(\pm\)0.0012 & 0.8181\fontsize{7}{7}\selectfont\(\pm\)0.0003 \\ 
        GAGA (WWW'23) & 0.9325\fontsize{7}{7}\selectfont\(\pm\)0.0044 & 0.9064\fontsize{7}{7}\selectfont\(\pm\)0.0082 & 0.8605\fontsize{7}{7}\selectfont\(\pm\)0.0038 \\
        \midrule
        \rowcolor{gray!10}
        \textbf{\shortname~(Ours)}  & \textbf{0.9694\fontsize{7}{7}\selectfont\(\pm\)0.0025} & \textbf{0.9534\fontsize{7}{7}\selectfont\(\pm\)0.0032} & \textbf{0.8949\fontsize{7}{7}\selectfont\(\pm\)0.0045} \\
        \bottomrule
    \end{tabular}}
\end{wraptable}

We also find that previous GNN-based methods perform extremely poorly compared to the Transformer-based method GAGA, even if the detectors are specifically designed.
This illustrates that fraud detection tasks primarily focus on modeling the semantic attributes of the target node.
However, the superiority of \shortname verifies that if topological information can be integrated into the node embedding with suitable fusion methods, the overall performance can be mutually enhanced instead of worsened.
Since we only use simple GCN as the topology modeling networks and do not make any effort to design networks oriented for the fraud detection task, GNN-based methods still have ample room for further improvement in how to cooperate with Transformer-based methods more effectively.

\subsection{Ablation Study}
\paragraph{Ablation on each submodule of \shortname}
To validate how each module of \shortname benefits the fraud detection performance, we conducted an ablation experiment to evaluate different components of \shortname. 
We present three baselines here: 1) GAGA, which is the Transformer-based state-of-the-art method; 2) Semantic Encoder, which removes the topology encoder from \shortname; and 3) Topology Encoder, which removes the semantic encoder from \shortname.
We experiment with all three baselines and \shortname on YelpChi and Pay datasets which possess more complex graphs than Amazon and amplify the differences between different methods.
As shown in Table~\ref{tab:ablation}, each component contributes to the proposed method. 
Semantic Encoder is slightly better than GAGA in all metrics on both datasets, which illustrates that gradually aggregating multi-relation information is beneficial for learning the semantic features. 
Using the topology encoder alone (Relation-Aware GCN) leads to a performance drop by 7.5\% in AUC, 28.0\% in AP, and 17.2\% in F1-macro on the YelpChi dataset, while it outperforms the semantic encoder on the larger dataset Pay in all metrics. 
This indicates that the relative importance of semantic features and topological features varies with the data. 
Among them, the topological features are more suitable in larger datasets, where the relation between nodes is more complex.
Moreover, the advocated \shortname, which integrates the topology encoder and the semantic encoder, always harvests the best performance in all metrics across both datasets. 
These results further emphasize the need to focus on both aspects of the graph. 
Our Relation-Aware GNN architecture can holistically capture topological information, making it an effective supplement to the semantic encoder, regardless of the size of the dataset.

\begin{figure}[tbp]
\begin{minipage}{0.5\textwidth}
    \captionof{table}{Ablation study on how the semantic encoder and topology encoder contribute to the proposed \shortname. }
    \label{tab:ablation}
    \resizebox{1.0\textwidth}{!}{
    \begin{tabular}{c|c|c|c|c|c|c}
        \toprule
        \multicolumn{1}{c|}{\multirow{2}{*}{\textbf{Methods}}} &\multicolumn{3}{c|}{\textbf{YelpChi}} &\multicolumn{3}{c}{\textbf{Pay}} \\
        \cmidrule{2-7} \multicolumn{1}{c|}{}& \textbf{AUC} & \textbf{AP} & \textbf{F1} & \textbf{AUC} & \textbf{AP} & \textbf{F1} \\
        \midrule
        GAGA & 0.9439 & 0.8014 & 0.8323 & 0.9325 & 0.9064 & 0.8605\\
        \midrule
        Semantic Encoder & 0.9454 & 0.8018 & 0.8343 & 0.9392 & 0.9120 & 0.8597 \\
        Topology Encoder & 0.9034 & 0.6420 & 0.7381 & 0.9500 & 0.9299 & 0.8653 \\
        \midrule
        \rowcolor{gray!10}
        \shortname  & \textbf{0.9781} & \textbf{0.9222} & \textbf{0.9098} & \textbf{0.9694} & \textbf{0.9534} & \textbf{0.8949} \\
        \bottomrule
    \end{tabular}}
\end{minipage}
\begin{minipage}{0.5\textwidth}
    \captionof{table}{Ablation study on fusion scheme. 
    We compare four other fusion methods. 
    Our method is highlighted with shading.
    }
    \label{tab:ablation-fusion}
    \resizebox{1.0\textwidth}{!}{
    \begin{tabular}{c|c|c|c|c|c|c}
        \toprule
        \multicolumn{1}{c|}{\multirow{2}{*}{\textbf{Fusion Scheme}}} &\multicolumn{3}{c|}{\textbf{YelpChi}} &\multicolumn{3}{c}{\textbf{Pay}} \\
        \cmidrule{2-7} \multicolumn{1}{c|}{}& \textbf{AUC} & \textbf{AP} & \textbf{F1} & \textbf{AUC} & \textbf{AP} & \textbf{F1} \\
        \midrule
       Concat & 0.9702 & 0.8834 & 0.8958 & 0.9612 & 0.9399 & 0.8927 \\
       Add & 0.9524 & 0.8075 & 0.8629 & 0.9592 & 0.9293 & 0.8737 \\
       Gated & 0.9653 & 0.8559 & 0.8898 & 0.9546 & 0.9049 & 0.8931 \\
       \midrule
        Attention w/o res  & 0.9720 & 0.8914 & 0.8995 & 0.9625 & 0.9375 & 0.8932 \\
       \rowcolor{gray!10}
        Attention w/ res  & \textbf{0.9781} & \textbf{0.9222} & \textbf{0.9098} & \textbf{0.9694} & \textbf{0.9534} & \textbf{0.8949} \\
        \bottomrule
    \end{tabular}
    }
\end{minipage}
\end{figure}

Apart from the superiority of our \shortname, it is also worth noting that the topology encoder has already surpassed many well-designed GNN-based fraud detectors, including CARE-GNN~\citep{dou2020enhancing}, PC-GNN~\citep{liu2021pick},  RioGNN~\citep{peng2021reinforced}, FRAUDRE~\citep{zhang2021fraudre} and H$^2$-FDetector~\citep{shi2022h2}. 
On YelpChi, Relation-Aware GCN can provide a performance gain of 31.1\% in AUC, 42.4\% in AP, and 23.1\% in F1-macro(compared to a single GCN). 
The embeddings learned from different relations can complement each other to a great extent. 
These results demonstrate that simply modeling each relation with a corresponding GNN in an end-to-end manner can lead to more efficient learning of the topological features.

\paragraph{Ablation on fusion modules}
There are a variety of effective methods to merge features from two embedding spaces.
Therefore, in this ablation study, we verify whether self-attention with residual connection is the first choice for the fusion of semantic and topological features in fraud detection task.
We consider four other feature fusion methods: (1) ``Concat'': directly concatenating the two features; (2) ``Add'': directly adding two feature vectors; (3) ``Gated'': using a learnable gate unit like GRU~\citep{cho-etal-2014-learning}. We use a linear layer followed by a sigmoid function to learn the weight $\gamma\in\mathbb{R}^{d}$ from $\boldsymbol{X}_{sem}$. In this setting, the final embedding $\boldsymbol{X}_{final}$ can be formulate as: $\boldsymbol{X}_{final} = \gamma*\boldsymbol{X}_{sem} + (1-\gamma)*\boldsymbol{X}_{gcn}$ , where $\boldsymbol{X}_{sem}$ represents the semantic features and $\boldsymbol{X}_{gcn}$ represents the topological features. 
(4) ``Attention w/ res'': a variant of our fusion module that removes the residual connection.
Similar to previous experiments, we use the same setting and apply the five methods on the YelpChi and Pay datasets.
We present the experiment results in Table~\ref{tab:ablation-fusion}.
It is evident that the direct concatenation and gate unit fall behind the self-attention but with a relatively smaller gap than simple addition.
However, these two methods lack the cross-modality handling ability that attention-based mechanisms have, thus leading to a suitable application of self-attention for fusing two features from different embedding spaces.
Furthermore, our experiments demonstrate that 'Attention w/ res' outperforms 'Attention w/o res' across all metrics, highlighting the effectiveness of residual connection in ensuring more robust fusion. In Addition, the residual connection can also accelerate the convergence of \shortname.
We present more ablation experiment results in Appendix~\ref{ablation-gnn} and~\ref{ablation-relation}

\subsection{Cost Analysis}
\label{cost}
\begin{wraptable}{r}{0.5\textwidth}
\vspace{-1.5em}
    \centering
    \caption{Cost Comparison on YelpChi. We quantify the resource consumption based on model parameters, MACs, and inference time. 
    }
    \label{tab:cost}
    \resizebox{0.5\textwidth}{!}{
    \begin{tabular}{c|c|c|c}
        \toprule
        \textbf{Methods/Module} & \textbf{Params(K)} & \textbf{MACs(M)} & \textbf{Inference Time(s)} \\
        \midrule
        GAGA & 64.61 & 2191.39 & 0.8963 \\
        \midrule
        Semantic Encoder & 73.92&2285.92 & /\\
        Topology Encoder &6.34 &33.47 & /\\
        Feature Fusion & 21.69 & 18.93 & /\\
        \midrule
        \shortname~(Ours)  & 101.95 & 2338.32 & 0.9984\\
        \bottomrule
    \end{tabular}}
\end{wraptable}

Previous methods either use GNN or Transformer to detect fraudulent nodes, while our \shortname combines GNN and Transformer together.
The integration of two sophisticated models inevitably leads to increased computational complexity and resource requirements.
Therefore, in this subsection, we quantify the resource consumption of each component of our \shortname and compare it with the state-of-the-art method GAGA~\citep{wang2023label}.

To comprehensively evaluate the resource consumption of our method, we measure it from three perspectives: the amount of model parameters, computation complexity (MACs), and inference time. The results are detailed in Table~\ref{tab:cost}. 
In comparison to GAGA, RAGFormer only exhibits a 6.7\% increase in computational overhead and an 11\% increase in inference latency. While there is a large increase in storage overhead, it is imperative to note that the absolute value of the storage volume remains minimal (expressed in kilobytes).   
It can be observed that the most resource-consuming component of \shortname is the Transformer-based semantic encoder. 
In contrast, the GNN-based topology encoder brings only a minor increase in resource consumption. 
Despite its significantly smaller size, the topology encoder can still capture the topological structure that the semantic encoder cannot learn, demonstrating the effectiveness of our Relation-Aware GNN design.

\begin{figure*}[tbp]
    \centering  
    \begin{subfigure}[c]{0.32\textwidth}
         \centering
        \includegraphics[width=\textwidth]{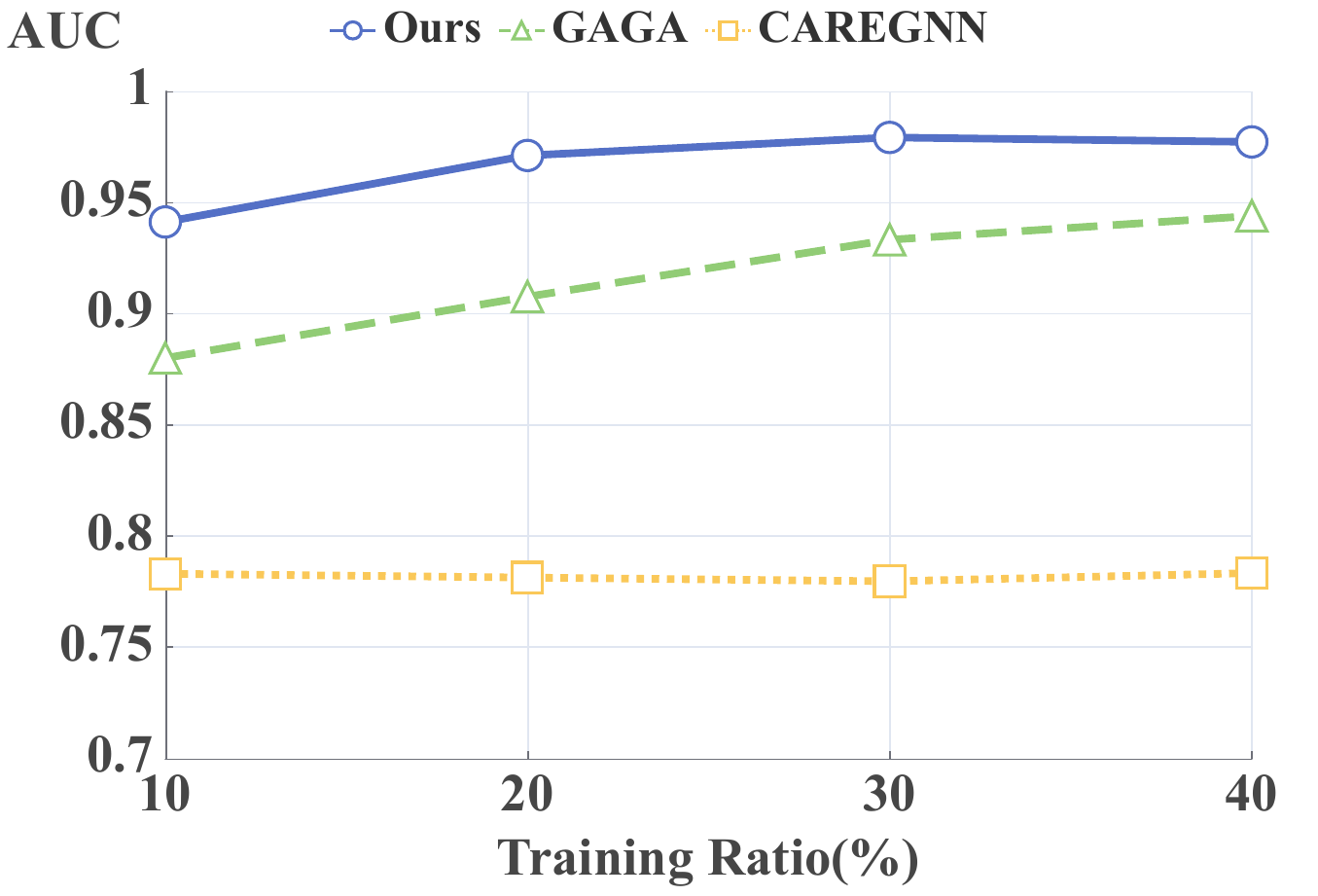}
         \caption{AUC on YelpChi
         }
         \label{fig:rate-auc}
     \end{subfigure}
     \begin{subfigure}[c]{0.32\textwidth}
         \centering
        \includegraphics[width=\textwidth]{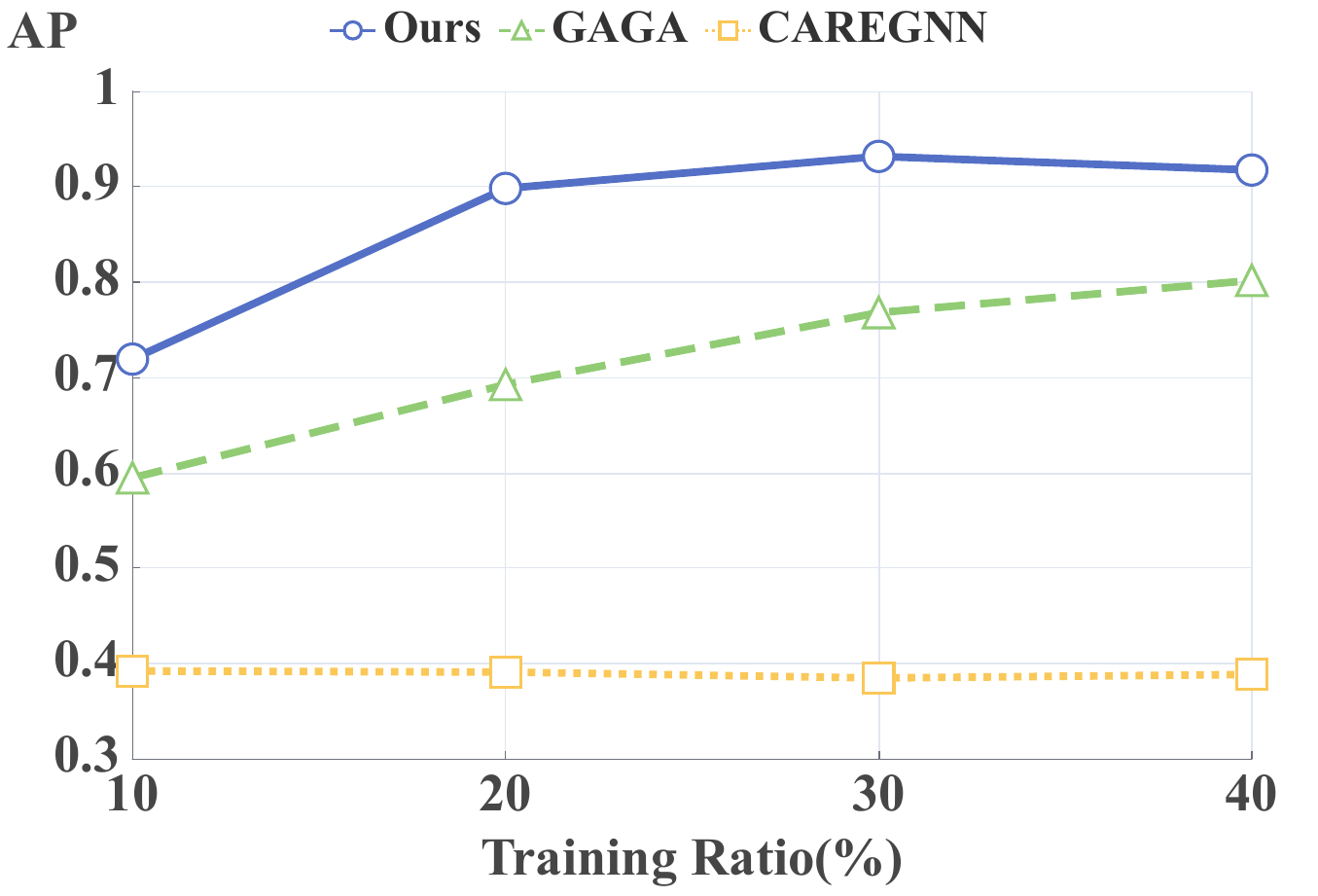}
         \caption{AP on YelpChi
         }
         \label{fig:rate-ap}
     \end{subfigure}
    \begin{subfigure}[c]{0.32\textwidth}
         \centering
        \includegraphics[width=\textwidth]{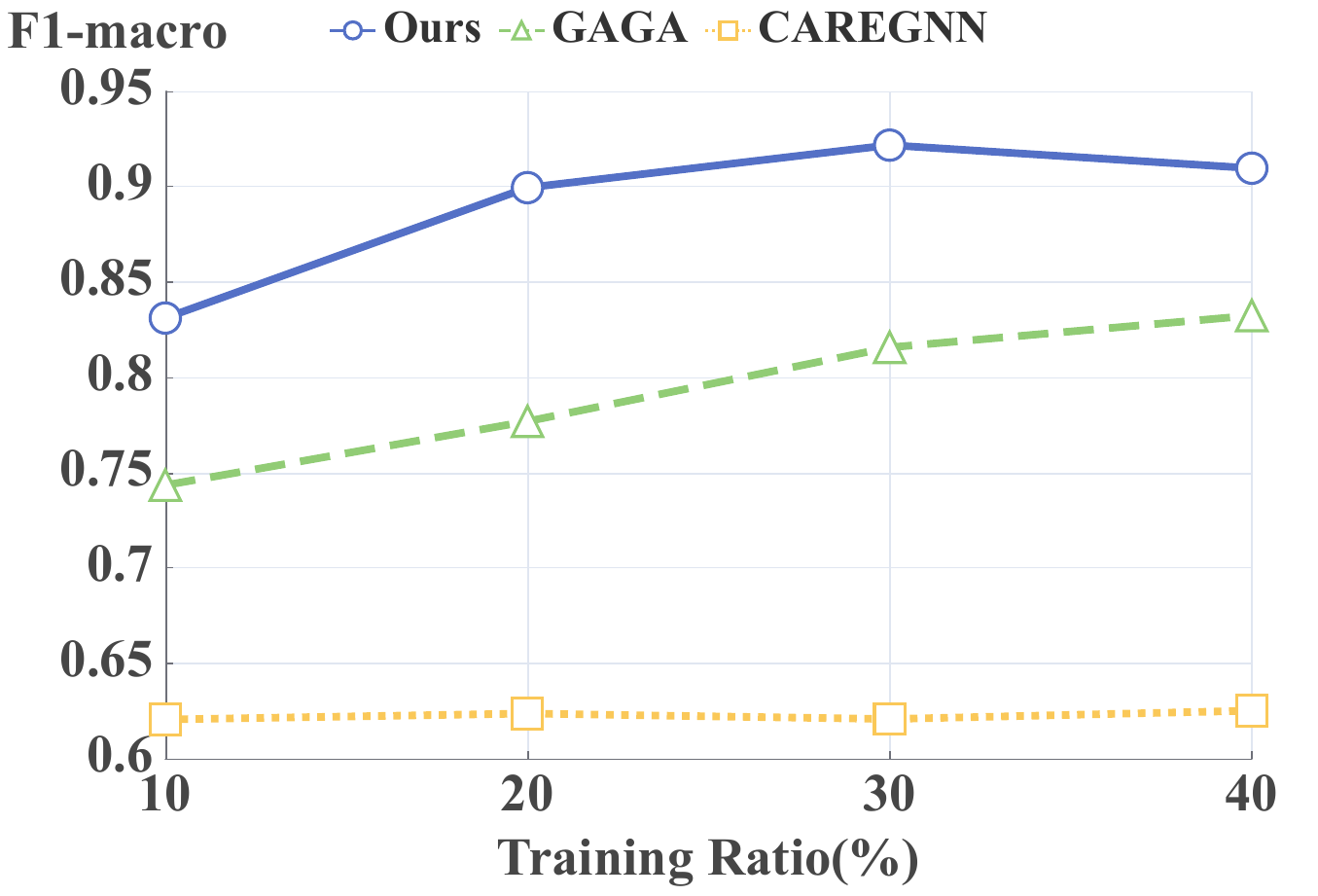}
         \caption{F1-macro on YelpChi
         }
         \label{fig:rate-f1}
     \end{subfigure}
\caption{The comparison of method performance of our \shortname, GAGA and CAREGNN on YelpChi with variable training ratios ranging from 10\% to 40\%. 
We evaluate all methods under AUC, AP, and F1-macro ((a), (b), (c)).
}
\label{fig:train-rate}
\end{figure*}
\subsection{Parameter Analysis}
\textbf{Training Ratio Analysis.}
In a semi-supervised setting, the proportion of the training set significantly influences the model's performance. To assess the robustness of the proposed algorithm in this context, we conducted experiments by varying the proportion of the training set.
Following previous works~\citep{dou2020enhancing, liu2021pick,wang2023label}, we set the proportions of the training set to 10\%, 20\%, 30\%, 40\% respectively. We compare the results of \shortname against GAGA and CAREGNN on the YelpChi dataset since CAREGNN is a widely used fraud detector with open-source code.
As depicted in Figure.~\ref{fig:train-rate}, \shortname exhibits remarkable performance across various training ratios, outperforming other methods by a substantial margin. 
In every setting, \shortname achieves the highest results. 
Notably, our method attains performance comparable to that of GAGA using merely 10\% of the training data, whereas GAGA requires 40\%.
This verifies that with the fusion of semantic and topological features, \shortname can make accurate judgments on node's belongings with much less data, which is highly advantageous for application on large graphs or in scenarios with limited annotated data. 
Furthermore, our approach attains its best performance at 30\% training ratio, indicating its remarkable data utilization capability. 
More results on Pay are provided in Appendix~\ref{train-ratio-app}.

\begin{figure}[tbp]
    \centering  
    \begin{subfigure}[c]{0.24\textwidth}
         \centering
        \includegraphics[width=\textwidth]{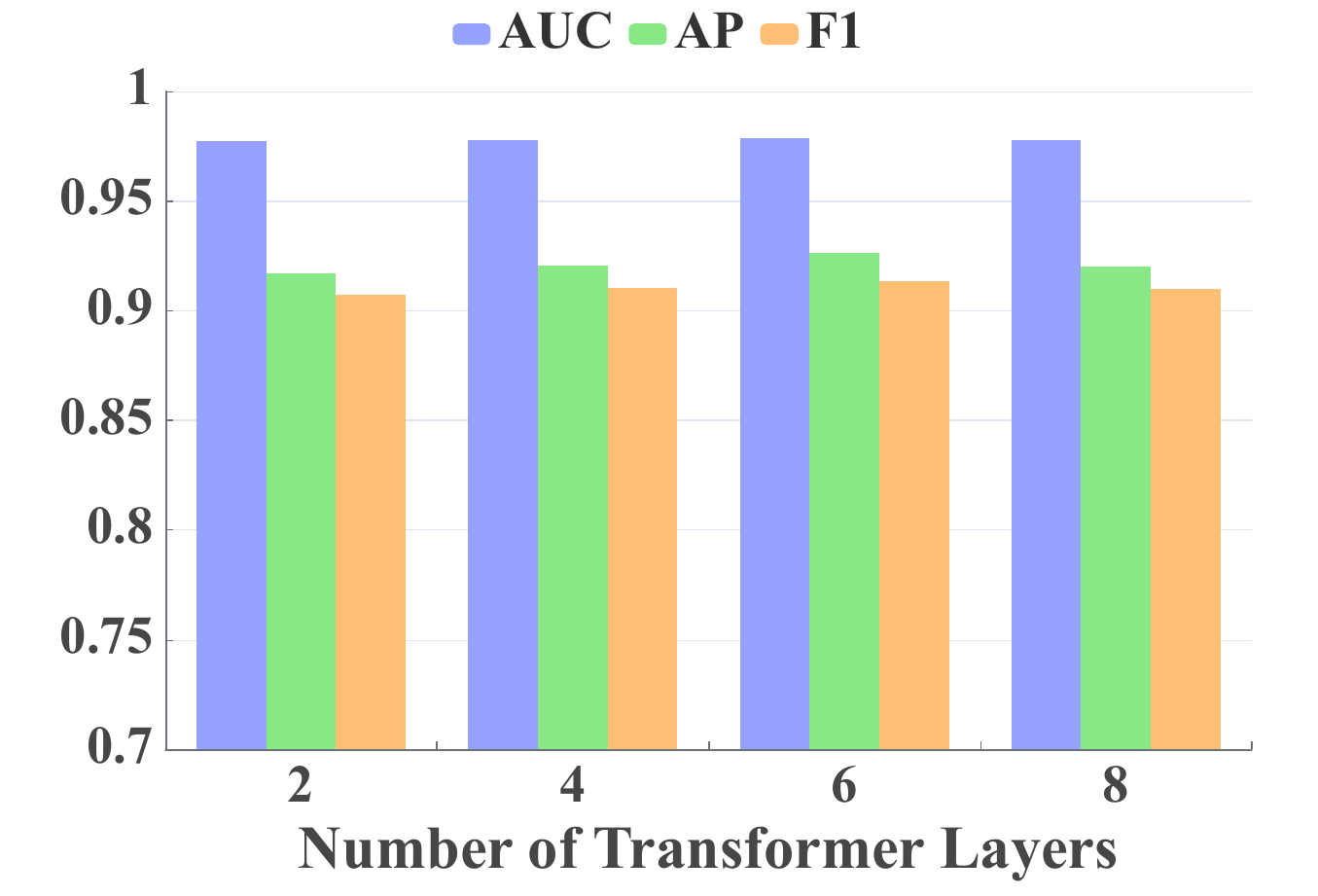}
        \caption{}
         \label{fig:hy-tl}
     \end{subfigure}
     \begin{subfigure}[c]{0.24\textwidth}
         \centering
        \includegraphics[width=\textwidth]{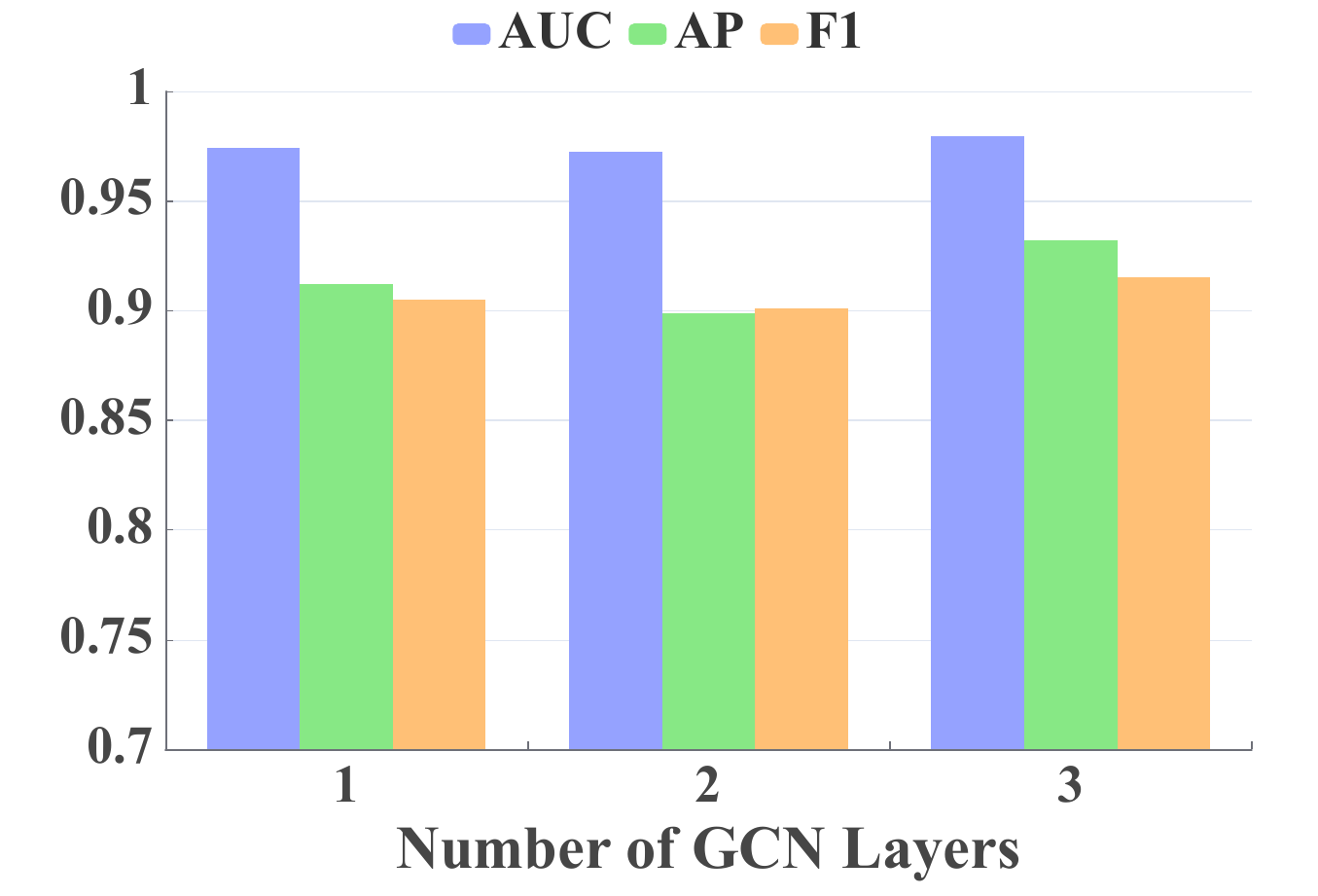}
        \caption{}
         \label{fig:hy-gl}
     \end{subfigure}
    \begin{subfigure}[c]{0.24\textwidth}
         \centering
        \includegraphics[width=\textwidth]{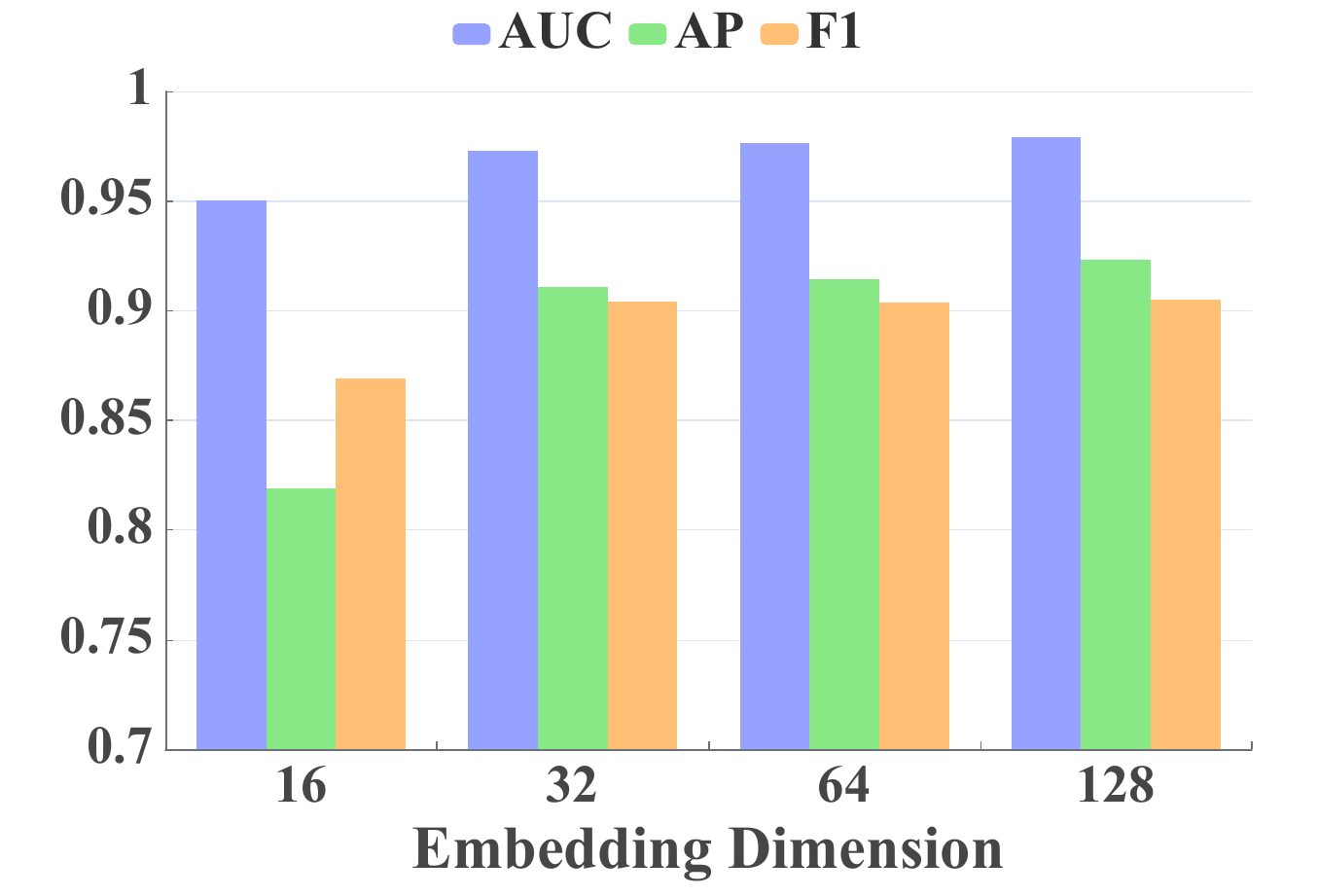}
        \caption{}
         \label{fig:hy-emb}
     \end{subfigure}
     \begin{subfigure}[c]{0.24\textwidth}
         \centering
        \includegraphics[width=\textwidth]{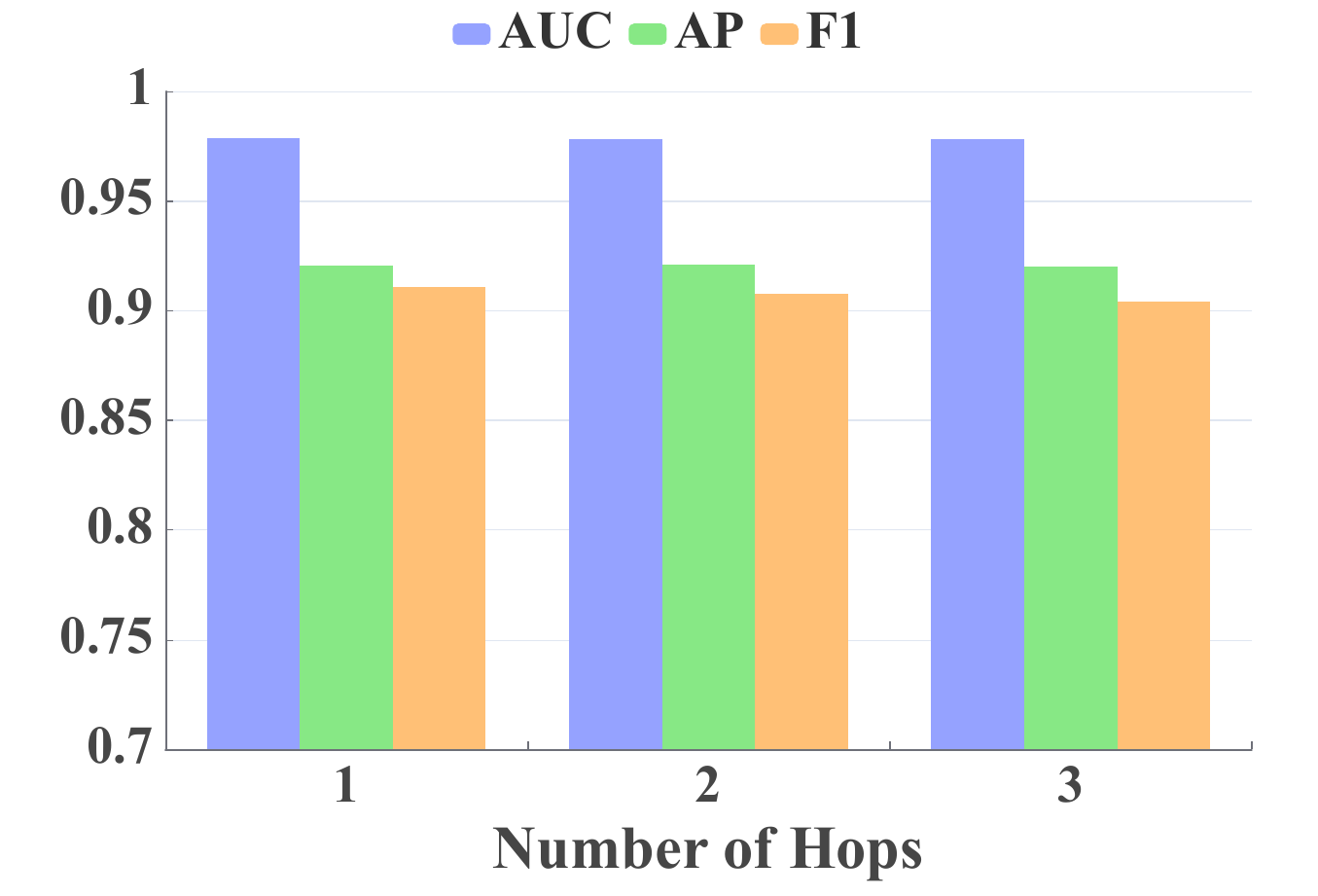}
        \caption{}
         \label{fig:hy-hop}
     \end{subfigure}
\caption{Hyper-parameter sensitivity analysis with respect to (a) the number of layers in the semantic encoder (b) the number of layers in the topology encoder (c) the embedding dimension of each relation and (d) the number of hops. All experiments are conducted on YelpChi.}
\label{fig:hyper}
\end{figure}

\textbf{Hyper-parameter Sensitivity Analysis.}
In order to investigate the sensitivity of \shortname against various hyper-parameters, we conducted additional evaluations on the YelpChi dataset. These evaluations focused on four key hyper-parameters: (1) the number of Transformer layers in the semantic encoder; (2) the number of GCN layers in the topology encoder; (3) the embedding dimension of each relation and (4) the number of hops in the semantic encoder.
These four hyper-parameters greatly affect the total number of parameters and thus impact the training cost and inference latency, while more parameters may bring consistent performance improvement. 
As evident from Figure~\ref{fig:hy-tl} and Figure~\ref{fig:hy-gl}, varying the depth of the semantic encoder or topology encoder does not affect model performance, which indicates that \shortname is highly flexible. 
Figure~\ref{fig:hy-hop} illustrates that increasing the number of hops has no substantial impact, which aligns with the findings of the preliminary experiment in Table~\ref{tab:preliminary}. 
In Figure~\ref{fig:hy-emb}, a low embedding dimension (16) results in noticeable performance degradation. 
When the embedding dimension reaches a sufficient magnitude (>=32), further increasing the embedding dimension has little impact on the performance of \shortname. 
This indicates that the model necessitates a sufficient embedding dimension to ensure an adequate representation capacity, but will not benefit from even larger embedding dimensions. 
Overall, the proposed method exhibits robustness and insensitivity to all four hyper-parameters.

\section{Conclusion}
In this study, we reveal that previous fraud detectors concentrate on learning only one aspect of multi-relation graphs, focusing either on the topology structure of the graph or the semantic attributes of individual nodes.
Based on this observation,
we propose a novel fraud detection method \shortname. 
Through attentively fusing semantic features and topological features into a unified framework,
\shortname is capable of capturing multiple perspectives of graph data, learning holistic node embeddings.
Extensive experiments on two benchmark opinion fraud datasets and an industrial financial fraud dataset demonstrate the effectiveness and robustness of our method.

{
\small

\bibliography{main}
}

\appendix
\newpage
\section*{\Large Appendix}
\section{Full Experiment Results}

\subsection{Detailed Datasets Description}
\label{dataset-app}
\begin{wraptable}{r}{0.5\textwidth}
    \centering
    \caption{Detailed statistics of YelpChi,  Amazon and Pay}
    \label{tab:datasets}
    \resizebox{0.5\textwidth}{!}{
    \begin{tabular}{c|ccc|cc}
        \hline
        Dataset & Nodes & Fraud\% & Feature Size & Relations & Edges \\
        \hline
        \multirow{3}{*}{YelpChi} & \multirow{3}{*}{45954} & \multirow{3}{*}{14.53\%} & \multirow{3}{*}{32} & R-U-R & 49,315 \\
        &    &     &  & R-T-R & 573,616 \\
           &     &     &  & R-S-R & 3,402,743 \\
        \hline
        \multirow{3}{*}{Amazon}  & \multirow{3}{*}{11944} & \multirow{3}{*}{9.50\%}  & \multirow{3}{*}{25} & U-P-U & 175,608 \\
                &       &         &    & U-S-U & 3,566,479 \\
                &       &         &    & U-V-U & 1,036,737 \\
        \hline
        \multirow{3}{*}{Pay}  & \multirow{3}{*}{64247} & \multirow{3}{*}{35.60\%}  & \multirow{3}{*}{147} & U-N-U & 9,954,549 \\
                &       &         &    & U-T-U & 17,754,106 \\
                &       &         &    & U-M-U & 17,727,314 \\
        \hline
    \end{tabular}}
\end{wraptable}

The YelpChi dataset comprises reviews of hotels and restaurants, categorizing filtered reviews as spam and recommended reviews as legitimate. It features three types of relations: R-U-R~(reviews posted by the same user), R-S-R~(reviews under the same product with the same rating), and R-T-R~(reviews under the same product posted in the same month). 
In terms of the Amazon dataset, it includes users' comments on musical instruments. 
Users with more than 80\% helpful votes are benign nodes while those with less than 20\% helpful votes are labeled as fraudulent. 
The Amazon dataset also has three relations: U-P-U~(users reviewing at least one same product), U-S-U~(users with at least one same star rating within one week), and U-V-U~(users with top-5\% mutual review similarities). 

To verify the effectiveness of \shortname in detecting financial fraud, we curated a real-world industrial dataset named Pay. Pay is designed for credit card fraud detection and comprises the transaction flow of users who used credit cards for transactions over a six-month period in 2023, along with detailed attributes of the corresponding users.
Those at a high risk of engaging in credit card arbitrage are labeled as fraudulent, whereas all others are considered benign. 
There exist three relations in Pay: 
U-N-U~(users transacting with the same merchant), U-T-U~(users transacting with merchants of the same category), and U-M-U~(users sharing the same M2 Delinquency Risk Scores for credit cards). 
We assure you that all user data utilized in our study has been anonymized to remove any personally identifiable information, ensuring privacy and compliance with ethical standards. 
Additionally, the data of Pay is securely stored and cannot be exported from our servers, safeguarding against unauthorized access or use. 
The detailed statistics of the three multi-relation datasets are shown in Table~\ref{tab:datasets}.

\subsection{Baselines}
We compare our method with three categories of methods as follows:
\begin{itemize}[leftmargin=*]
\item \textbf{Classical GNN model}: We choose several general GNN models that capture structural information, including GCN~\citep{kipf2016semi}, GAT~\citep{velivckovic2017graph}, GraphSAGE~\citep{hamilton2017inductive}and GraphSAINT~\citep{zeng2019graphsaint}. We also implement heterogeneous model HAN~\citep{han2019} as a baseline. 
\item \textbf{GNN-based fraud detector}: GraphConsis~\citep{liu2020alleviating} tried to tackle the context, feature and relation inconsistency problem in a multi-relation graph. CARE-GNN~\citep{dou2020enhancing} and PC-GNN~\citep{liu2021pick} utilized node or neighbor filtering to mitigate the class imbalance issue.  RioGNN~\citep{peng2021reinforced} proposed a reinforcement learning framework to choose similar neighbors of a target node. FRAUDRE~\citep{zhang2021fraudre}, and COFRAUD~\citep{zang2023don} explored novel approaches to model multi-relation interactions.  H$^2$-FDetector~\citep{shi2022h2} learned both homophilic and heterophilic connections between the nodes. 
\item \textbf{Transformer-based fraud detector}: GAGA~\citep{wang2023label} is a state-of-the-art method that employs Transformer to better capture semantic information.
\end{itemize}

\subsection{Preliminary Experiment on Feature Similarity}
\label{pre-app}
\begin{figure}[tbp]
    \centering  
    \begin{subfigure}[c]{0.48\textwidth}
         \centering
        \includegraphics[width=\textwidth]{Styles/figures/yelpchi_cosine_similarity.pdf}
        \caption{Cosine Similarity on YelpChi}
         \label{fig:sim-y-cos}
     \end{subfigure}
     \begin{subfigure}[c]{0.48\textwidth}
         \centering
        \includegraphics[width=\textwidth]{Styles/figures/yelpchi_centered_kernel_alignment.pdf}
        \caption{CKA on YelpChi}
         \label{fig:sim-y-cka}
     \end{subfigure}
    \begin{subfigure}[c]{0.48\textwidth}
         \centering
        \includegraphics[width=\textwidth]{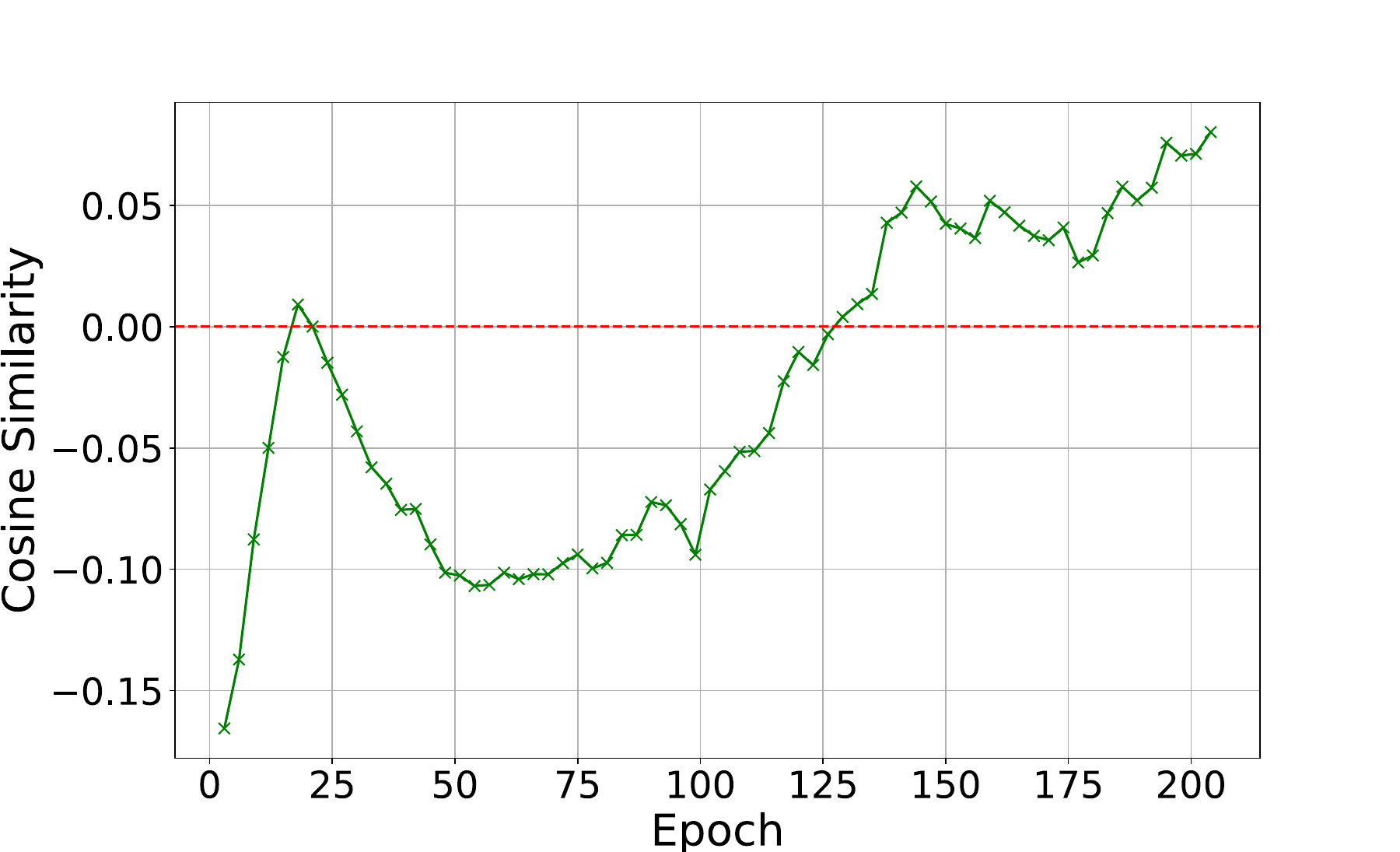}
        \caption{Cosine Similarity on Amazon}
         \label{fig:sim-a-cos}
     \end{subfigure}
     \begin{subfigure}[c]{0.48\textwidth}
         \centering
        \includegraphics[width=\textwidth]{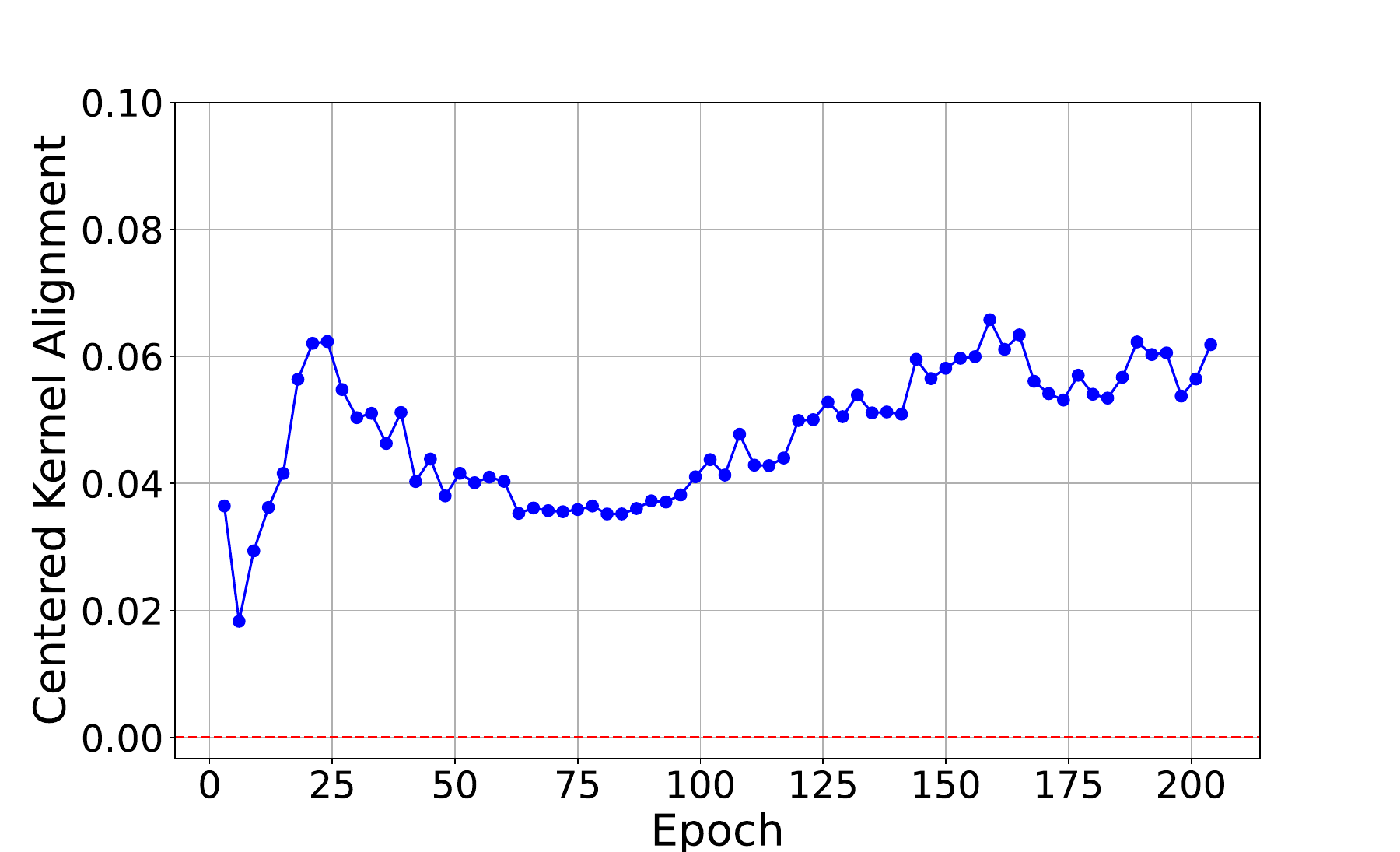}
        \caption{CKA on Amazon}
         \label{fig:sim-a-cka}
     \end{subfigure}
\caption{Similarity between features learned by our semantic encoder(Transformer) and topology encoder(GNN). 
A value of 0, indicated by the red line, signifies that the features are orthogonal.}
\label{fig:sim}
\end{figure}
To validate our intuition, we evaluate the similarity between features learned by our semantic encoder and topology encoder on two datasets, YelpChi~\citep{rayana2015collective} and Amazon~\citep{mcauley2013amateurs}. We adopt cosine similarity and Centered Kernel Alignment (CKA)~\citep{kornblith2019similarity} as metrics and illustrate their changes over training epochs in Fig.~\ref{fig:sim}. 
CKA measures the similarity between two sets of features, with values close to 1 indicating strong similarity and values near 0 indicating little or no similarity. 
On YelpChi, as the number of training epochs increases, both the cosine similarity and CKA score between the semantic features and topology features gradually approach zero, indicating that the features are nearly orthogonal with low correlation. 
On Amazon, fluctuations in the two similarity metrics are more pronounced, yet they still maintain low values(<0.08). 
The representations learned by the semantic encoder and the topology encoder are nearly orthogonal, yet both are effective for fraud detection. 
This confirms our intuition that Transformer and GNN indeed focus on different aspects of the graph, thereby suggesting that combining the two could further boost performance.

\subsection{Ablation on GNN Backbone}
\label{ablation-gnn}

\begin{minipage}{1.0\textwidth}
    \captionof{table}{Ablation study on GNN backbones. We select GraphSAGE and GAT as two additional baselines as these two backbones are not well-designed fraud detectors.}
    \label{tab:ablation-gnn}
    \resizebox{1.0\textwidth}{!}{
    \begin{tabular}{c|c|c|c|c|c|c}
        \toprule
        \multicolumn{1}{c|}{\multirow{2}{*}{\textbf{GNN backbone}}} &\multicolumn{3}{c|}{\textbf{YelpChi}} &\multicolumn{3}{c}{\textbf{Pay}} \\
        \cmidrule{2-7} \multicolumn{1}{c|}{}& \textbf{AUC} & \textbf{AP} & \textbf{F1} & \textbf{AUC} & \textbf{AP} & \textbf{F1} \\
        \midrule
       GraphSAGE & 0.9723 & 0.9081 & 0.9068 & 0.9369 & 0.9144 & 0.8672 \\
       GAT & 0.9074 & 0.7245 & 0.8389 & 0.9603 & 0.9406 & 0.8888 \\
       \midrule
       \rowcolor{gray!10}
       GCN~(ours)  & \textbf{0.9781} & \textbf{0.9222} & \textbf{0.9098} & \textbf{0.9694} & \textbf{0.9534} & \textbf{0.8949} \\
        \bottomrule
    \end{tabular}}
\end{minipage}

\paragraph{Abaltion on GNN types}
To explain why choosing GCN as the backbone of the Relation-Aware GNN in \shortname, we replaced GCN with GraphSAGE and GAT to validate whether GCN is the best choice for modeling topological features for supplement of the semantic encoder.
Following the same setting in \shortname, we use two-layer GraphSAGE or GAT as the backbone of the topology encoder and leave the rest of the settings unchanged, which results in a fair comparison with GCN in terms of parameter counts.
Specifically, we use a mean aggregator for GraphSAGE and we follow DGL~\cite{wang2019dgl} to set the attention dropout and feature dropout rate both to 0.6 and set the number of heads in multi-head attention as eight in GAT.
From Table~\ref{tab:ablation-gnn}, we can observe that 1) using GraphSAGE/GAT as the backbone of the Relation-Aware GNN outperforms single GraphSAGE/GAT and many GNN-based fraud detectors. Specifically, Relation-Aware GraphSAGE performs better on YelpChi, while Relation-Aware GAT is more effective on Pay.
2) Both GNN methods are less superior to GCN used in our \shortname, which features a simpler design than those two methods.
This indicates that our method is highly efficient and effective.

\subsection{Ablation on the contribution of each relation graph}
\label{ablation-relation}

\begin{wraptable}{r}{0.5\textwidth}
    \centering
    \caption{Ablation study on relation graph.}
    \label{tab:ablation-relation}
    \resizebox{0.5\textwidth}{!}{
    \begin{tabular}{c|c|c|c}
        \toprule
        \multicolumn{1}{c|}{\multirow{2}{*}{\textbf{Relation}}} &\multicolumn{3}{c|}{\textbf{YelpChi}} \\
        \cmidrule{2-4} \multicolumn{1}{c|}{}& \textbf{AUC} & \textbf{AP} & \textbf{F1} \\
        \midrule
       R-U-R & 0.8740 & 0.5734 & 0.7356 \\
       R-T-R & 0.9123 & 0.6871 & 0.7586 \\
       R-S-R & 0.9585 & 0.8394 & 0.8358 \\
       \midrule
       R-U-R\&R-T-R & 0.9400 & 0.7848 & 0.8307 \\
       R-U-R\&R-S-R & 0.9738 & 0.8970 & 0.8993 \\
       R-T-R\&R-S-R & 0.9685 & 0.8796 & 0.8798 \\
       \midrule
       \rowcolor{gray!10}
       All & \textbf{0.9781} & \textbf{0.9222} & \textbf{0.9098}\\
        \bottomrule
    \end{tabular}}
\end{wraptable}

\paragraph{Abaltion on relation graph}
Fraud detection methods focus on extracting information from multi-relation graphs. To assess the validity of each relation graph, we conducted an ablation experiment on the YelpChi dataset to determine the contribution of each individual relation. 
The YelpChi dataset includes three types of relations, resulting in a total of six subgraphs: three single-relation graphs and three dual-relation graphs.
We applied \shortname on all six subgraphs as part of our ablation experiments. When only one relationship is present, the Relation-Aware GCN degenerates into a single GCN.
The results of \shortname on these subgraphs are shown in Table~\ref{tab:ablation-relation}.
Our findings are as follows:
\textbf{(1)} The results indicate that utilizing a single relation or a pair of relations yields inferior performance compared to leveraging all three relations simultaneously. Furthermore, using two relations concurrently outperforms using a single relation. This demonstrates that the three relations effectively complement one another and that incorporating all three relations is the most efficient approach.
\textbf{(2)} Each time an additional relation is incorporated, the performance shows a notable improvement. 
This indicates that our \shortname can effectively leverage multiple relations, making it well-suited for learning on complex multi-relation graphs.
\textbf{(3)} Even with only one relation (R-S-R), our \shortname can significantly outperform the state-of-the-art method GAGA, which fully proves the advantage of learning both semantic features and topological features.
\textbf{(4)} Among the three relations, the R-S-R relation is the most effective. Combining R-S-R with another relation can achieve performance nearly equivalent to those obtained by using all three relations.

\begin{figure*}[tbp]
    \centering  
    \begin{subfigure}[c]{0.32\textwidth}
         \centering
        \includegraphics[width=\textwidth]{Styles/figures/AUC.pdf}
         \caption{AUC on YelpChi
         }
         \label{fig:rate-auc}
     \end{subfigure}
     \begin{subfigure}[c]{0.32\textwidth}
         \centering
        \includegraphics[width=\textwidth]{Styles/figures/AP.pdf}
         \caption{AP on YelpChi
         }
         \label{fig:rate-ap}
     \end{subfigure}
    \begin{subfigure}[c]{0.32\textwidth}
         \centering
        \includegraphics[width=\textwidth]{Styles/figures/F1-macro.pdf}
         \caption{F1-macro on YelpChi
         }
         \label{fig:rate-f1}
     \end{subfigure}
     \begin{subfigure}[c]{0.32\textwidth}
         \centering
        \includegraphics[width=\textwidth]{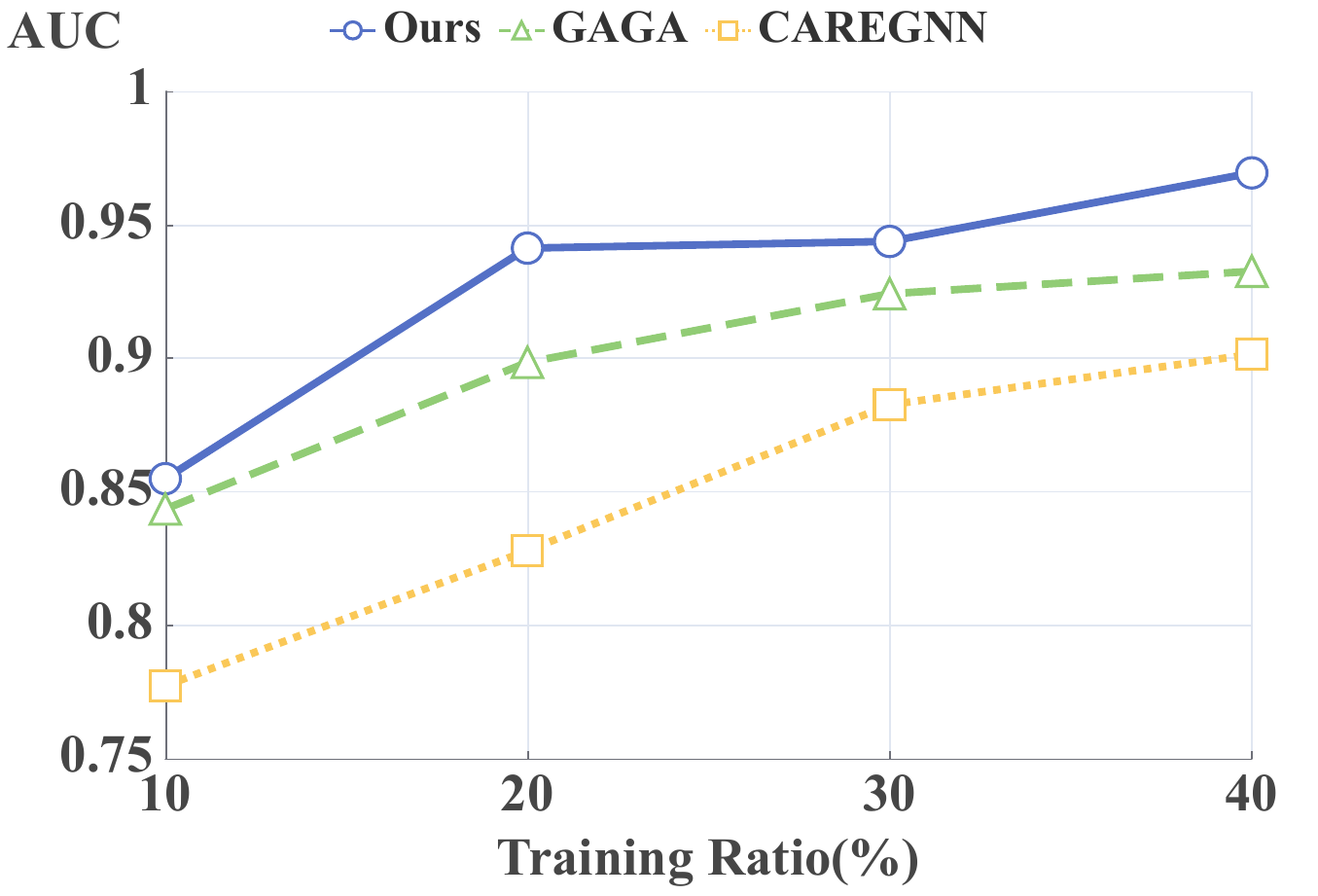}
         \caption{AUC on Pay
         }
         \label{fig:rate-auc-c}
     \end{subfigure}
     \begin{subfigure}[c]{0.32\textwidth}
         \centering
        \includegraphics[width=\textwidth]{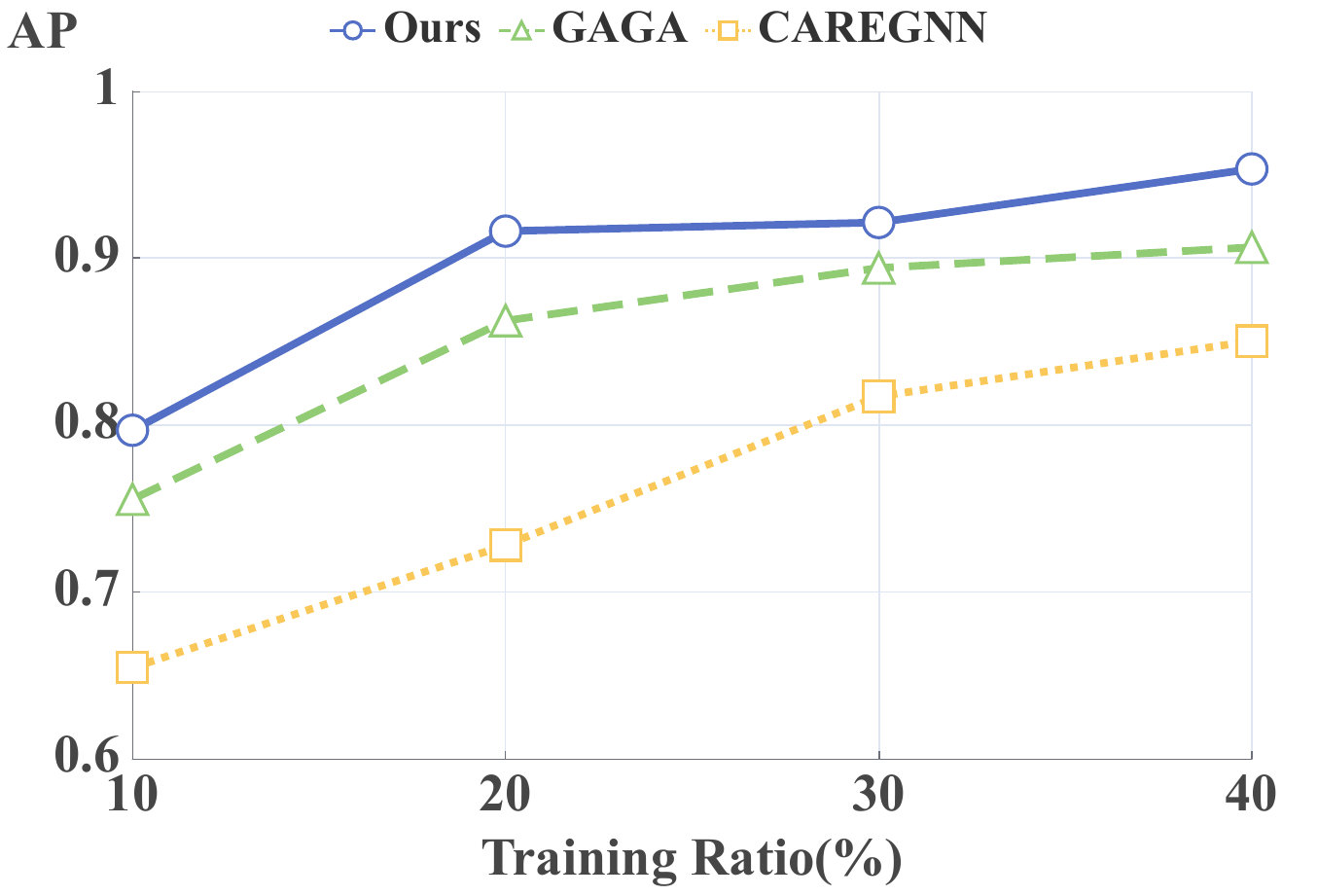}
         \caption{AP on Pay
         }
         \label{fig:rate-ap-c}
     \end{subfigure}
     \begin{subfigure}[c]{0.32\textwidth}
         \centering
        \includegraphics[width=\textwidth]{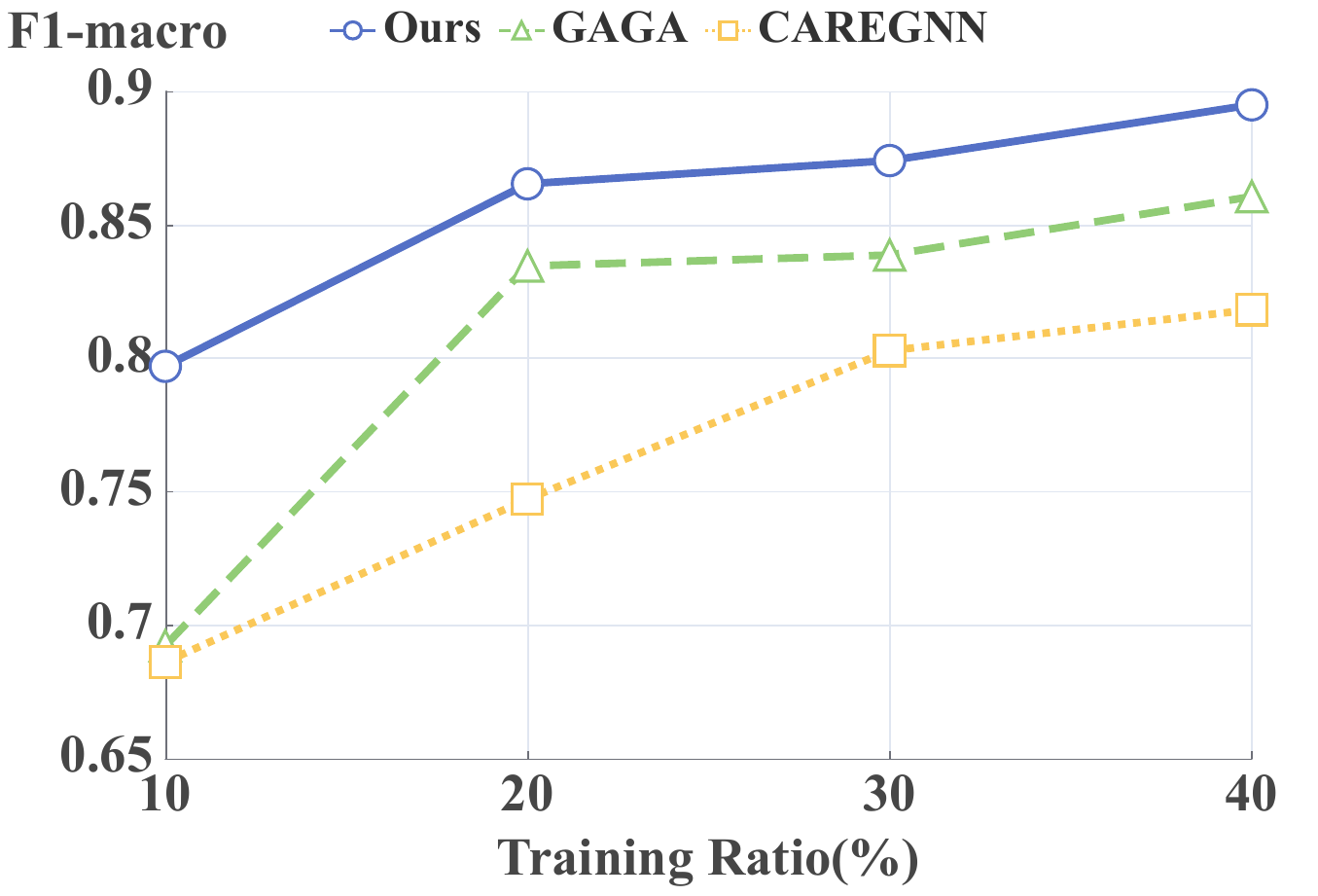}
         \caption{F1-macro on Pay
         }
         \label{fig:rate-f1-c}
     \end{subfigure}
\caption{The comparison of method performance of our \shortname, GAGA and CAREGNN on two datasets with variable training ratios ranging from 10\% to 40\%. We evaluate all methods under three metrics (AUC, AP, and F1-macro) on YelpChi ((a), (b), (c)) and Pay ((d), (e), (f)), respectively.}
\label{fig:train-rate}
\end{figure*}

\subsection{Training Ratio Analysis}
\label{train-ratio-app}
We present the full results of the training ratio analysis, including results on Pay.

In a semi-supervised setting, the proportion of the training set significantly influences the model's performance. To assess the robustness of the proposed algorithm in this context, we conducted experiments by varying the proportion of the training set.
Following previous works~\citep{dou2020enhancing, liu2021pick,wang2023label}, we set the proportions of the training set to 10\%, 20\%, 30\%, 40\% respectively. We compare the results of \shortname against GAGA and CAREGNN on YelpChi dataset since CAREGNN is a widely used fraud detector with open-source code.
As depicted in Figure~\ref{fig:train-rate}, \shortname exhibits remarkable performance across various training ratios, outperforming other methods by a substantial margin. 
In every setting, \shortname achieves the highest results in all metrics. 
Notably, our method attains performance comparable to that of GAGA using merely 10\% of the training data, whereas GAGA requires 40\%.
This verifies that with the fusion of semantic and topological features, \shortname can make accurate judgments on node's belongings with much less data, which is highly advantageous for application on large graphs or in scenarios with limited annotated data. 
Furthermore, our approach attains its best performance at 30\% training ratio, indicating its remarkable data utilization capability. 
On the Pay dataset, \shortname performs best with only 20\% training data across the other two baselines using double training data, which also indicates that \shortname is not hungry for data and converges fast.

\section{Further Discussions}
\subsection{Broader Impact}
\label{ssec:broader_impact}
In this paper, we reviewed previous fraud detection methods from the perspective of feature similarity and found that the features learned by GNN-based models and Transformer-based models are highly orthogonal. 
Through detailed experimentation, we discovered a critical limitation in existing methods: they predominantly focus on learning only one aspect of the multi-relation graph.
To tackle this issue, we propose the \shortname framework for learning both semantic and topological features on multi-relation graphs.
The proposed method not only contributes to the technical field of fraud detection but also promotes a more secure socio-economic environment.
Our objective is to build stronger fraud detection method to prevent fraudulent activities, thereby safeguarding financial resources and enhancing the reliability of financial transactions.
We also acknowledge the importance of considering ethical implications to prevent potential misuse of this technology.

\subsection{Limitations}
The proposed method employs a Transformer-based semantic encoder and a GNN-based topology encoder to learn node features. 
The integration of two sophisticated models, while beneficial, inevitably leads to increased computational complexity and resource demands.
Compared to previous methods, our \shortname requires more storage and computing resources.
A detailed comparison can be found in~\ref{cost}.
Such requirements may limit its applicability in resource-constrained environments or in systems where rapid processing is critical. 
Furthermore, the current validation of our \shortname has been confined to opinion fraud and credit card fraud. 
To ensure the robustness and applicability of our method, it is imperative to conduct further tests across a broader range of fraud types. 

\end{document}